\documentclass[a4paper]{article}

\usepackage{amsfonts}
\usepackage{amsmath}
\usepackage{graphicx}
\usepackage{pslatex}

\def\DF#1{\textbf{#1}}

\def\FORALL{\bigwedge}
\def\EXISTS{\bigvee}

\def\Z{\mathbb{Z}}

\def\R{\mathbb{R}}

\def\P{\mathcal{P}}
\def\c{\mathcal{C}}

\def\T{\mathcal{T}}

\def\SUBSET{\subseteq}

\def\THEN{\Rightarrow}
\def\IF{\Rightarrow}
\def\ONLYIF{\Leftarrow}

\def\C{\mbox{\textnormal{C}}}
\def\dim{\mbox{\textnormal{dim}}}

\def\FORALL{\bigwedge}
\def\EXISTS{\bigvee}

\newtheorem{Satz}{Theorem}
\newtheorem{Def}{Definition}
\newtheorem{Lemma}{Lemma}
\newtheorem{Folg}{Corollary}

\newenvironment{Beweis}
{\paragraph{\textit{Proof.}}}
{\hspace{\stretch{1}}$\Box$\vspace{5mm}}
\def\qed{\hspace{\stretch{1}}$\Box$}

\def\COMP#1{{#1}^c}
\def\SETMINUS{\setminus}

\title{
Digital Manifolds and the Theorem of Jordan-Brouwer
}

\author{Martin H\"unniger}
\date{ }

\begin{document}
\maketitle

\pagestyle{plain}
\pagenumbering{roman}

\begin{abstract}
  We give an answer to the question given by T.Y.Kong in his
  article \emph{Can 3-D Digital Topology be Based on Axiomatically
    Defined Digital Spaces?} \cite{kong}. In this article he asks the
  question, if so called ``good pairs'' of neighborhood
  relations can be found on the set $\Z^n$ such that the existence of
  digital manifolds of dimension $n-1$, that separate their complement in
  exactly two connected sets, is guaranteed. 
  To achieve this, we use a technique developed by M. Khachan
  et.al. \cite{khachan}. A set given in $\Z^n$ is translated into a
  simplicial complex that can be used to study the topological
  properties of the original discrete point-set. In this way, one is
  able to define the notion of a  $(n-1)$-dimensional digital manifold
  and prove the digital analog of the Jordan-Brouwer-Theorem.
\end{abstract}
\tableofcontents

\pagenumbering{arabic}


\section{Introduction}\label{chapter_5}

Since the beginning of the research in digital image processing, the
question of the definition of a sound topological framework
arose. Though in the two dimensional case a solution was easy to find,
its generalization to higher dimensions seemed very hard. This is easy
to see from the vast amount of theories under consideration by the
community. The central goal was to find preconditions so that a 
discrete analog to the theorem of Jordan-Brouwer--the generalization 
of the Jordan curve theorem to arbitrary dimensions--is satisfied. 
The Jordan curve theorem states that every closed curve in the plane, 
that is simple, i.e. has no crossings, separates the plane in exactly 
two regions: Its inside and its outside and is itself the boundary of 
both of these sets.

\begin{figure}[htb]
  \begin{center}
    \includegraphics{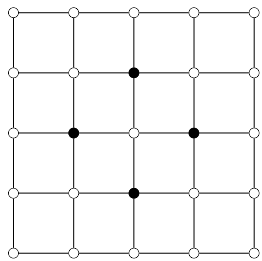}
  \end{center}
\caption[Initial problem]{\small Depending on the adjacency relations we use for the 
black and the white points, respectively, the set of black points is connected 
(8-adjacency) or disconnected (4-adjacancy). Also the set of white points may be connected
(8-adjacency) or disconnected (4-adjacency). Only 4-adjacency is depicted.}
\label{Pic:ausgangs_problem}
\end{figure}

In this paper, we restrict the discrete setting to lattices $\Z^n$ for $n\ge 2$. 
This is sufficient, since it is possible to embed other settings in these sets
and $\Z^n$ is very suitable in geometric terms as Albrecht H\"ubler shows for the 
2-dimensional case \cite{huebler}. The main focus is on the adjacency relations
with which we add structure to the $\Z^n$. As figure \ref{Pic:ausgangs_problem} 
shows, the validity of a discrete ``Jordan curve'' theorem depends on the adjacency relations
we apply to the points. It is in general not sufficient to use the same adjacencies
for the white and black points, i.e. the background and the foreground, as this 
picture shows and so we have to deal with pairs of such relations. The pairs of 
adjacencies that make it possible to define a correct discrete notion of a simple 
closed curve are called ``good pairs''. In this paper we will solve the problem
of defining a discrete $(n-1)$-manifold in $\Z^n$, which is the $n$-dimensional
analog to a simple closed curve, and characterize the good pairs in all dimensions
greater than 2.

The first person to give an idea for the 2-dimensional case was A.~Rosenfeld 
\cite{rosenfeld} in 1973. E.~Khalimsky \cite{khalimsky} studied very
special topological adjacency relations, which were suitable in any
dimension, during the early 1980s. In the 1990s, G.T.~Herman \cite{herman}
proposed a framework with very general neighborhood relations, he
tried to make the Jordan curve property to a property of pairs of
points. Unfortunately the approach does not resemble the intuition
given by the Euclidean case. Anyway, the theory showed a promising
method, how to generalize the concept of a good pair to higher
dimensions. Later, in 2003, M.~Khachan et al. \cite{khachan} brought
together the theory of pairs of the form $(2n,3^{n}-1)$ in arbitrary
dimensions $n$. They were the first ones using the notion of the
simplicial complex, a basic structure known from algebraic
topology. This approach led to deeper insights, but it was bound to
the very special adjacency-relations used by the authors.  

T.Y.~Kong \cite{kong} renewed the question of finding a general
theory for the problem of topologization of $\Z^n$ in 2001. 
The Approach we a taking is a new definition of good pairs in dimensions 
greater than 2 and a new general concept of a $(n-1)$-manifold in $\Z^n$. 
For a given $(n-1)$-manifold $M$ in $\Z^n$ we are able to construct a
simplicial complex $K(M)$, that preserves the topological properties of $M$, 
if $\Z^n$ is endowed with a good pair of adjacencies. 
The complex $K(M)$ then has nice topological
properties--it is a so called Pseudomanifold (as studied by P.S.~Alexandrov 
\cite{alexandrov}) and therefore, we are able to embed it in $n$-dimensional 
Euclidean space in a natural way. By doing this we are defining a real manifold 
if and only if the adjacencies on $\Z^n$ were chosen correctly and the real
version of the Jordan-Brouwer Theorem can be applied.

The paper is organized as follows: In section 2 we give basic
definitions to make precise the topological and graph theoretic
terms. In section 3 we give the definition of a digital manifold and 
study its basic properties. Also we give a new and general definition of a
good pair. In section 4 we introduce the Theorem of
Jordan-Brouwer. Together with the notions of digital manifolds and
good pairs, we are able to construct a simplicial complex with the same
topological properties as the digital manifold. Having this tools established,
we can give the general proof of the discrete variant of
the Theorem of Jordan-Brouwer in arbitrary dimensions in section
5. The proof consists of three parts, all involving heavy case
differentiations of technical nature. We end the paper with
Conclusions in section 6. 

\section{Basic Definitions}

First of all, we review some definitions from the field of simplicial
complexes. Our goal is to construct these objects from subsets of
$\Z^n$. 

\subsection{Simplices and Simplicial Complexes}
\begin{Def}
Let $x_0,\ldots,x_k\in\R^n$ be affine independent
points. The set 
\begin{equation} 
  \sigma=\{x\in\R^n: x=\sum_{i=0}^k\lambda_ix_i\mbox{ with
  }\sum_{i=0}^k\lambda_i=1,\lambda_0,\ldots,\lambda_q> 0\}\SUBSET\R^n
\end{equation}
is the (open) \DF{simplex with vertices} $x_0,\ldots,x_k$. We also
write $\sigma=(x_0,\ldots,x_k)$. The number $k$ is the dimension
of $\sigma$. Sometimes, for brevity, we call $\sigma$ just $k$-Simplex.
\end{Def}

Let $\sigma,\tau\SUBSET\R^n$ be simplices, then $\tau$ is called
\DF{face} of $\sigma$, in terms: $\tau\le\sigma$, if the vertices of
$\tau$ are also vertices of $\sigma$. The relation $\tau<\sigma$ means
$\tau\le\sigma$ and $\tau\neq\sigma$.

\begin{Def}\label{simplizialkomplex}
  A \DF{simplicial complex} $K$ in $\R^n$ is a finite set of simplices
  in $\R^n$ with the following properties
  \begin{enumerate}
  \item For any $\sigma\in K$ and $\tau<\sigma$ is $\tau\in K$.
  \item For any $\sigma,\tau\in K$ with $\sigma\neq\tau$ holds $\sigma\cap\tau=\emptyset$
  \end{enumerate}
\end{Def}

\label{geometrische_realisierung}
  Let $K$ be a simplicial complex in $\R^n$. The geometric realization
  of $K$ is the set 
\begin{equation}
  \bigcup_{\sigma\in K}\sigma\SUBSET\R^n\enspace .
\end{equation}

A simplicial complex $K$ is \DF{homogenous $(n-1)$-dimensional}, if
every simplex in $K$ is face of a $(n-1)$-simplex in $K$. A homogenous
$(n-1)$-dimensional simplicial complex $K$ is \DF{strongly
  connected}, if for any two $(n-1)$-simplices $\sigma$ and $\sigma'$
exists a sequence of $(n-1)$-simplices
$\sigma=\sigma_0,\ldots,\sigma_l=\sigma'$, such that $\sigma_i$
and $\sigma_{i+1}$ share a common $(n-2)$-face for every
$i\in\{0,\ldots,l-1\}$. 
The complex $K$ is \DF{non-degenerated}, if every $(n-2)$-simplex $\sigma$ is face of
exactly two $(n-1)$-simplices.

\begin{Def}\label{pseudomannigfaltigkeit}
  A simplicial complex $K$ is a \DF{combinatorial
    $(n-1)$-pseudomanifold} without border, if
  \begin{enumerate}
  \item $K$ is homogenous $(n-1)$-dimensional,
  \item $K$ is non-degenerated,
  \item $K$ is strongly connected.
  \end{enumerate}
\end{Def}

\subsection{Adjacencies}

To establish structure on the points of the set $\Z^n$ we have to
define some kind of connectivity relation.

\begin{Def}\label{adjazenzrelation}
  Given a set $\P$, a relation  $\alpha\SUBSET\P\times\P$ is called \DF{adjacency}
  if it has the following properties:
  \begin{enumerate}
  \item $\alpha$ is finitary: $\forall p\in\P:|\alpha(p)|<\infty$.
  \item $\P$ is connected under $\alpha$.
  \item Every finite subset  of $\P$ has at most one infinite connected component 
    as complement.
  \end{enumerate}
\end{Def}

A set $M\SUBSET\P$ is called \DF{connected} if for any two points 
$p,q$ in $M$ exist points $p_0,\ldots,p_m$ and a positive integer $m$ such that
$p_0=p$, $p_m=q$ and $p_{i+1}\in A(p_i)$ for all
$i\in\{0,\ldots,m-1\}$. 

In the text we will consider pairs $(\alpha,\beta)$ of adjacencies
on the set $\Z^n$. In this pair $\alpha$ represents the adjacency
on a set $M\SUBSET\Z^n$, while $\beta$ represents the adjacency on 
$\COMP{M}=\Z^n\SETMINUS M$.

Let $\T$ be the set of all translations on the set $\Z^n$.
The generators $\tau_1,\ldots,\tau_n\in\T$ of  $\Z^n$
induce a adjacency $\pi$ in a natural way: 

\begin{Def} Two points $p,q$ of $\Z^n$ are called \DF{proto-adjacent}, in terms
  $p\in\pi(q)$, if there exists a $i\in\{1,\ldots,n\}$ such that
  $p=\tau_i(q)$ or $p=\tau_i^{-1}(q)$. 
\end{Def}

We can use the standard base of $\R^n$ as the generators of $\Z^n$, since all sums of 
integer multiples of this base is a point in $\Z^n$.

Another important adjacency on $\Z^n$ is \DF{$\omega$}.
\begin{equation}
  \omega(p):=\{q\in\Z: |p_i-q_i|\le 1, 0\le i\le n\}  
\end{equation}

\begin{Lemma}
  For every $n\ge 2$ and all $p\in\Z^n$ the set $\omega(p)$ is connected 
  under $\pi$. \qed
\end{Lemma}

In the rest of the text let $\alpha$ and $\beta$ be two adjacencies on 
$\Z^n$ such that for any $p\in\Z^n$ holds
\begin{equation}
  \pi(p)\SUBSET\alpha(p),\beta(p)\SUBSET\omega(p)\enspace.
\end{equation}

\begin{Lemma}
  The set $\Z^n$ is connected under $\pi$.
\end{Lemma}

\begin{Beweis}
  Let  $p,q$ be any two points in $\Z^n$. We need to show that there exists a
  $\pi$-path from $p$ to $q$. Let $p=(p_1,\ldots,p_n)$ and
  $q=(q_1,\ldots,q_n)$. We prove by induction on
  $k=\sum_{j=1}^n|q_i-p_i|$. In the case $k=1$ the points $p$ and $q$ only differ
  in one coordinate $i$ by 1, since all terms in the sum are positive and integral. 
  It follows $\tau_i(p)=q$ respectively $\tau_i^{-1}(p)=q$ and thus $q\in\pi(p)$. 
  In the case $k>1$ we look for the smallest index $i\in\{1,\ldots,n\}$ such that $p_i\neq q_i$. 
  The point
  \begin{equation}
    p'=\left\{
    \begin{array}{rcl}
      (p_1,\ldots,p_i+1,\ldots,p_n) &\mbox{if}& q_i>p_i \\
      (p_1,\ldots,p_i-1,\ldots,p_n) &\mbox{if}& q_i<p_i \\
    \end{array}
    \right.
  \end{equation}
  is by definition $\pi$-adjacent to $p$ and 
  \begin{equation}
    \sum_{j=1}^n|q_j-p_j'| = \sum_{j=1,j\neq i}^n|q_j-p_j| +
    |q_i-p_i|-1
  \end{equation}
  By the induction hypothesis exists a $\pi$-path
  \begin{equation}
    p'=p^{(0)},\ldots,p^{(k-1)}=q
  \end{equation}
  and $p,p^{(0)},\ldots,p^{(k-1)}$ is the path we are looking for.
\end{Beweis}

\begin{Def}
  A point $p\in M\SUBSET\Z^n$ is called \DF{simple} for the pair
  $(\alpha,\beta)$ if the following properties hold:
  \begin{enumerate}
  \item $M$ and $M\SETMINUS\{p\}$ contain the same number of
    $\alpha$-connected components.
  \item $\COMP{M}$ and $\COMP{(M\SETMINUS\{p\})}$ contain the same number of
    $\beta$-connected components.
  \end{enumerate}
\end{Def}

\begin{figure}[htb]
  \begin{center}
    \includegraphics[scale=1.0]{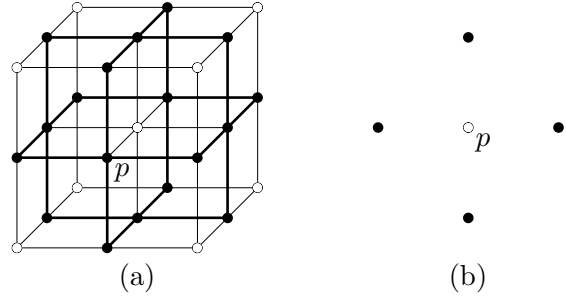}
  \end{center}
  \caption[Graph-theoretic surfaces don't need to be topological surfaces]
          {\small{(a) The depicted set is connected under $\pi$
              and could be understood as digital surface. (b) The $\pi$-neighborhood
              of a point $p$ is not necessarily homotopic to a curve as the topological
              case would imply.}} 
          \label{Pic:graph_nicht_top_flaeche}
\end{figure}

\subsection{The Separation Property}

We call the set 
\begin{equation}
C^k=\{0,1\}^k\times\{0\}^{n-k}\SUBSET\Z^n
\end{equation}
the $k$-dimensional standard \DF{cube} in $\Z^n$.
The set $C^k$ can be embedded in ${n}\choose{k}$ different ways in $C^n$.
A general $k$-cube in $\Z^n$ is defined by a translation of a standard cube.

Indeed, we can construct any $k$-cube $C$ from one point $p$ with $k$ generators
in the following way:
\begin{equation}
  C = \{\tau_1^{e_1}\cdot\tau_k^{e_k}(p): e_i\in\{0,1\}, i=1,\ldots,k \}
\end{equation}
The dimension of $C'$ is then $k+l$. We use this construction in the next definition.

\begin{Def}\label{trenndef}
  Let $M\SUBSET \Z^n$, $n \ge 2$ and $C$ be a $k$-cube,
  $2\le k\le n$. The complement of $M$ is in $C$ \DF{not separated} by $M$ under
  the pair $(\alpha,\beta)$, if 
  for every $\alpha$-component $M'$ of $C\cap M$ and every
  $(k-2)$-subcube $C^*$ of $C$ the following is true:
  
  If $C^*$ is such that $C^*\cap M'\neq\emptyset$ has maximal cardinality among 
  all sets of this form, and the sets $\tau_1(C^*)\SETMINUS M$ and 
  $\tau_2(C^*)\SETMINUS M$ are both nonempty and lie in one common $\beta$-component 
  of $\COMP{M}$, then holds
  \begin{equation}   
    (\tau_1\tau_2)^{-1}(\tau_1\tau_2(C^*)\cap M') \SUBSET
    \tau_1^{-1}(\tau_1(C^*)\cap M') \cap
    \tau_2^{-1}(\tau_2(C^*)\cap M')
    \enspace .
  \end{equation}
\end{Def}

\begin{figure}[htb]
  \begin{center}
    \includegraphics[scale=1.0]{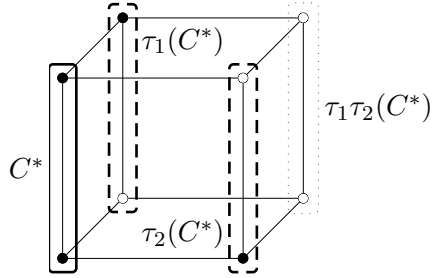}
  \end{center}
  \caption[The separation property]{\small{$C^*$ has an intersection
      of maximal cardinality with the $\alpha$-component
      $M'$. The sets $\tau_1(C^*)\SETMINUS M'$ and $\tau_2(C^*)\SETMINUS M'$ are
      nonempty and belong to a $\beta$-component of
      $\COMP{M}$. Since $\tau_1\tau_2(C^*)\cap M' = \emptyset$,
      the property of definition \ref{trenndef} is satisfied for this $C^*$.
      But the set $M'$ separates $\COMP{M}$ in the cube $C^*$. Why?}}  
  \label{Pic:trennungseigenschaft1}
\end{figure}

In the following, we only consider the case when $C\cap M$ has at most one
$\alpha$-component. This can be justified by viewing any other
$\alpha$-component besides the one considered as part of the
background, since there is no $\alpha$-connection anyway.
This property also gets important if we study the construction of the
simplicial complex.

A set $M$ has the \DF{separation property} under a pair $(\alpha,\beta)$, 
if for every $k$-cube $C$, $2\le k\le n$ as in the definition \ref{trenndef} 
the set $\COMP{M}$ is in $C$ not separated by $M$

\begin{figure}[htb]
  \begin{center}
    \includegraphics[scale=1.0]{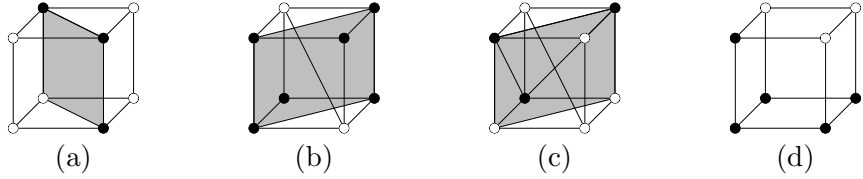}
  \end{center}
  \caption[The separation property 2]{\small{The black points represent the set $M$ in the
      given 3-cubes. The white points represent the complement of $M$. In the cases (a) to (c)
      the complement, which is connected, is separated by $M$. This separation is depicted 
      by the gray plane spanned by $C^*$ and $\tau_1\tau_2(C^*)$. 
      In Figure (d) occurs no separation, 
      since the only choice for $C^*$ would be a 1-cube, that contains only black 
      points. 
  }} 
  \label{Pic:trennungseigenschaft}
\end{figure}

The meaning of the separation property is depicted in the figure
\ref{Pic:trennungseigenschaft}.

\begin{Lemma}\label{komponentenschranken}
  Let $C$ be a $k$-cube for $0\le k\le n$ that has exactly two $\beta'$-components
  in $C\SETMINUS M$. Then we can bound the number of points in $\C\cup M$ as follows:
  \begin{equation}
    k\le |C\cup M| \le 2^k-2
  \end{equation}
\end{Lemma}

\begin{Beweis}
  The upper bound follows from the existence of at least two points in $C\SETMINUS M$ 
  and the observation that a $k$-cube has $2^k$ vertices.

  The lower bound stems from the case that one of the two $\beta$-components is a 
  singleton. Because then at least al the neighbors of this one point need to be 
  in $M$. There are $k$ such points.
\end{Beweis}

\begin{Lemma}\label{punkte-anzahl}
  A $\beta$-component of $C\SETMINUS M$ in a $k$-cube $C$ with $l$ elements
  has at least
  \begin{equation}
    (k-m)\cdot l+ 2^m-l,\qquad m=\lceil\log_2{l}\rceil
  \end{equation}
  $\beta$-neighbors in $M$. 
\end{Lemma}

\begin{Beweis}
  First, we fill a $m$-cube $C^m$ with the $l$ points such that $m$ is a 
  minimal integer and $C^m$ contains all the points. This cube can be translated
  in $C$ in $(k-m)$ ways. Therefore, each of the $l$ points in $C^m$ has 
  $(k-m)$ $\pi$-neighbors in $C$. Finally, we can also have $2^m-l$ neighbors in $C^m$.
\end{Beweis}

\begin{Lemma}\label{dimensionskonsistenz}
  Let $M\SUBSET\Z^n$ be a set that has the separation property and for every
  cube $C$ let $M\cap C$ be $\alpha$-connected. Let $K$ be any $\beta$-component 
  of $C\SUBSET M$. Then in every subcube $C'\SUBSET C$  at most one 
  $\beta$-component of $C'\SETMINUS M$ is contained in $K$.
\end{Lemma}

This lemma guarantees that one $\beta$-components of any $C\SETMINUS M$ cannot be split
into two or more components in any subcube of $C$.

\begin{Beweis}
  Assume for contradiction that $C$ is a $k$-cube that contains a $l$-subcube $C'$ 
  such that two $\beta$-com\-ponents of $C'\SETMINUS M$ lie in one $\beta$-component 
  of $C\SETMINUS M$. In $C'\SETMINUS M$ are at least two $\beta$-components.
  For any $(l-2)$-subcube $C^*$ of $C'$ exist certain generators $\tau_1$ 
  and $\tau_2$ such that
  \begin{equation}
    C' =C^*\cup\tau_1(C^*)\cup\tau_2(C^*)\cup\tau_1\tau_2'(C^*)
  \end{equation}
  If $C^*$ and $\tau_1(C^*)$ lie completely in $M$ then we consider instead of the 
  given $C'$ the new $(l-1)$-cube $C'=\tau_2(C^*)\cup\tau_1\tau_2(C^*)$. 

  Starting with the case $l=2$ we can see, that there exists a $(l-2)$-subcube
  $C^*$ of $C'$ that contains a maximal number of points of $M$, otherwise, 
  $C'\SETMINUS M$ would contain only one $\beta$-component.

  Using the separation property for this $C^*$ we find
  \begin{equation}
    (\tau_1\tau_2)^{-1}(\tau_1\tau_2(C^*)\cap
    M)\SUBSET(\tau_1^{-1}(\tau_1(C^*)\cap
    M))\cap(\tau_2^{-1}(\tau_2(C^*)\cap M))
  \end{equation}
  This shows that $C'\SETMINUS M$ is $\pi$-connected. This contradicts our assumption
  on the existence of two $\beta$-components in $C'$.
\end{Beweis}


\section{Digital Manifolds}

\begin{Def}\label{n-1-mannigfaltigkeit}
  An $\alpha$-connected set $M\SUBSET\Z^n$, for $n\ge 2$, is a
  \DF{(digital) $(n-1)$-manifold} under the pair
  $(\alpha,\beta)$, if the following properties hold:
  \begin{enumerate}
  \item\label{cubeconnection} In any $n$-cube $C$ the set $C\cap M$
    is $\alpha$-connected.
  \item\label{twocomponents} For every $p\in M$ the set
    $\omega(p)\SETMINUS M$ has exactly two $\beta$-components
    $C_p$ and $D_p$. 
  \item\label{componentunity} For every $p\in M$ and every
    $q\in\alpha(p)\cap M$ the point $q$ is $\beta$-adjacent to $C_p$ and $D_p$.
  \item\label{separationproperty} $M$ has the separation property.
  \end{enumerate}
\end{Def}

\begin{figure}[htb]
  \begin{center}
    \includegraphics[scale=1.0]{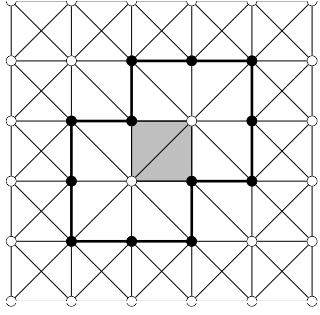}
  \end{center}
  \caption{\small{The independence of the property \ref{cubeconnection} from the other
      properties of the $(n-1)$-manifold. The black points of the gray cube are 
      not $\alpha$-connected, while the other properties are satisfied.}}  
  \label{Pic:unabhaengigkeit_1}
\end{figure}

\begin{figure}[htb]
  \begin{center}
    \includegraphics[scale=1.0]{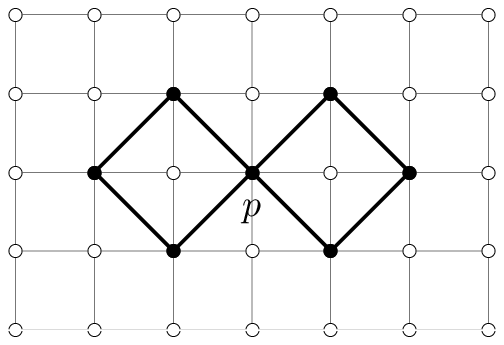}
  \end{center}
  \caption{\small{The independence of property \ref{twocomponents} from the
      other properties of a $(n-1)$-manifold. For the point $p$ the set
      $\omega(p)\SETMINUS M$ has four $\beta$-components. }} 
  \label{Pic:unabhaengigkeit_2}
\end{figure}

\begin{figure}[htb]
  \begin{center}
    \includegraphics[scale=1.0]{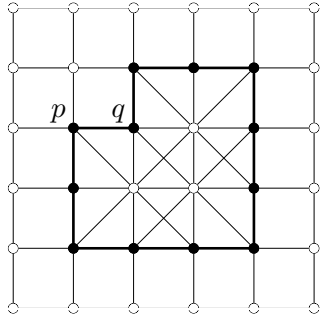}
  \end{center}
  \caption{\small{The independence of property \ref{componentunity} from the
      other properties of a $(n-1)$-manifold. The point $q$ is in $\omega(p)$
      $\beta$-adjacent to only one $\beta$-component of $\omega(p)\SETMINUS M$.}} 
  \label{Pic:unabhaengigkeit_3}
\end{figure}

\begin{figure}[htb]
  \begin{center}
    \includegraphics[scale=1.0]{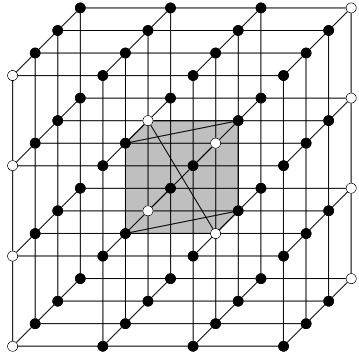}
  \end{center}
  \caption{\small{The independence of the property \ref{separationproperty} from the
      other properties of a $(n-1)$-manifold. To satisfy the other properties, adjacencies 
      of points can be easily inserted. They have not been depicted for clarity.
      The gray 3-cube violates the separation property.
      Also compare this situation to Figure \ref{Pic:trennungseigenschaft}.(a).}} 
  \label{Pic:unabhaengigkeit_4}
\end{figure}

The independence of the four properties is depicted in the figures
\ref{Pic:unabhaengigkeit_1} to \ref{Pic:unabhaengigkeit_4}.

\begin{Satz}\label{satz-zwei-komp}
  Let $M\SUBSET\Z^n$, $n\ge 2$, be a $(n-1)$-manifold under the pair $(\alpha,\beta)$. 
  Then the set $\omega(M)\SETMINUS M$ contains exactly two $\beta$-components.
\end{Satz}

\begin{Beweis}
  We know:
  \begin{equation}
    \omega(M)=\{q\in\Z^n:\EXISTS_{p\in M}:q\in\omega(p)\}
  \end{equation}
  Let $p$ be any point in $M$. The set $\omega(q)\SETMINUS M$ consists
  per definitionem of the two $\beta$-components $C_p$ and
  $D_p$. Every point $q\in\alpha(p)$ is $\beta$-adjacent to both of these
  components. Furthermore, $p$ is $\beta$-adjacent to both of them.
  The set $\omega(q)$ also consists of two $\beta$-components $C_q$ and $D_q$,
  whereby the naming can be made consistent by saying $C_q$ is the component 
  that contains a common point with $C_p$ and analogue for $D_q$.

  $C_p$ and $C_q$ are nonempty and contain a common point, therefore the union 
  of the both sets is $\beta$-connected.

  Let $a$ and $b$ be any two points of $\omega(M)\SETMINUS M$. By definition
  there exist two points $p$ and $q$ with $a\in\omega(p)$ and $b\in\omega(q)$.
  Since $M$ is $\alpha$-connected, there exist an $\alpha$-path $P$ of the
  form $p=p_{0},p_{1},\ldots,p_{k}=q$ in $M$. This path induces the
  following sets that are $\beta$ connected by the observation above:
  \begin{equation}
    C_P:=\bigcup_{i=0}^k C_{p^{(i)}}
  \end{equation}
  and
  \begin{equation}
    D_P:=\bigcup_{i=0}^k D_{p^{(i)}}
  \end{equation}

  If $a$ and $b$ lie in the same set $C_P$ or $D_P$ respectively, then
  the two points are $\beta$-connected. Otherwise no $\beta$-path 
  between the two sets can exist.
\end{Beweis}

Let $M\SUBSET\Z^n$ be a $(n-1)$-manifold.
The following two definitions base upon the last theorem:
\begin{equation}
  C_M:=\bigcup\{C_P:\mbox{for every path $P$ between any two points
  $p,q\in M$}\}
\end{equation}
\begin{equation}
  D_M:=\bigcup\{D_P:\mbox{for every path $P$ between any two points
  $p,q\in M$}\}
\end{equation}
We recapitulate the notion of the elementary equivalence of paths given 
by G.T. Herman \cite{herman} Let $w$ and $w'$ be paths in $\P$ under the 
adjacency $\alpha$ of the form
\begin{eqnarray}
  w(0),\ldots,w(l),&w(l+1),\ldots, w(l+n-1),&w(l+k),\ldots,w(l+k+m) \\
  w'(0),\ldots,w'(l),&w'(l+1),\ldots,w'(l+k-1),&w'(l+n),\ldots,w'(l+n+m)
\end{eqnarray}
that satisfy $w(i)=w'(i)$, for $i\in\{0,\ldots,l\}$, $w(l+n+i)=w'(l+k+i)$, for 
$i\in\{0\ldots,m\}$. Then $w$ and $w'$ are called \DF{elementary
  $N$-equivalent}, if $1\le k+n\le N+2$.

Two paths $w$, $w'$ in $\P$ are called \DF{$N$-equivalent}, if there
is a sequence $w_0,\ldots,w_p$ of $\alpha$-paths with $w_0=w$ and
$w_p=w'$, such that $w_{i-1}$ and $w_i$ are elementary $N$-equivalent
for $i\in\{1\ldots,p\}$. Spaces, in which every cycle is $N$-equivalent to
a point, are called \DF{$N$-simple connected}.

A $(n-1)$-manifold $M\SUBSET\Z^n$, $n\ge 2$, under the pair $(\alpha,\beta)$ is a 
\DF{$(n-1)$-sphere}, if it is $N$-simple connected for some positive
integer $N$.

A pair $(\alpha,\beta)$ of adjacency relations on $\Z^n$ is a
\DF{separating pair} if for all $p\in\Z^n$ the set $\beta(p)$ is a
$(n-1)$-sphere.

\subsection{Double Points}\label{doppelpunkt}
\begin{Def}
  A point $p\in\beta(z)$ $z\in\Z^n$ is a \DF{double point} under the
  pair $(\alpha,\beta)$, if there exist points
  $q\in\pi(z)\cap\alpha(p)$ and $r\in\beta(z)\cap\pi(p)$ 
  and a simple\footnote{A translation $\tau$  is called \DF{simple} if no other translation
  $\sigma$ exists with $\sigma^n=\tau$, $n\in\Z, |n|\neq 1$. } 
  translation $\tau\in\T$ with $\tau(p)=q$, $\tau(r)=z$ and $q\in\alpha(r)$. 
\end{Def}

This concept is the key to a local characterization of the good pairs 
$(\alpha,\beta)$. Without it, one could not consistently define topological
invariants like the Euler-char\-acter\-istic. It means that an edge
between points in a set $M$ can be crossed by an edge between points
of its complement and these four points lie in a square defined by the
corresponding adjacencies. This crossing can be seen as a double point,
belonging both to the foreground \emph{and} to the background. Also,
mention the close relationship to the separation property, which is a
more general concept of similar interpretation. 

\begin{figure}[htb]
\begin{center}
\includegraphics{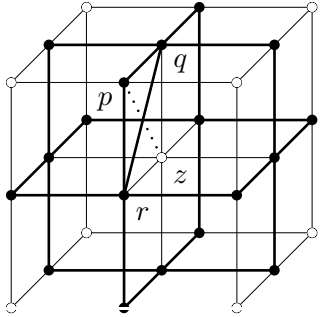}
\end{center}
\caption[A double point]{\small{A double point $p$. The fat edges
    represent the $\alpha$-adjacency. Only the relevant edges have been drawn for
    clarity. The dotted edge represents the $\beta$-adjacency of $p$ and $z$. The 
    black points are the $\beta$-neighbors of $z$.}} 
\label{Pic:doppelpunkt}
\end{figure}

An adjacency relation $\alpha$ on $\Z^n$ is \DF{regular}, if at least one
of the following holds:
\begin{enumerate}
\item For every translation $\tau$ on $\Z^n$ is $p\in\alpha(q)$,
  iff $\tau(p)\in\alpha(\tau(q))$.
\item For every rotation $\varrho$ on $\Z^n$ is $p\in\alpha(q)$,
  iff $\varrho(p)\in\alpha(\varrho(q))$. 
\end{enumerate}

\subsection{Degenerations}

\begin{Lemma}\label{the_prop}
  Let $(\alpha,\beta)$ be a separating pair of regular adjacencies
  on $\Z^n$. If $\beta(p)$ contains a double point for any $p\in\Z^n$,
  then exists a set, for which the Euler-characteristic cannot be
  defined.
\end{Lemma}

\begin{Beweis}
  Our argumentation follows section 7.5.3 in Klette and
  Rosenfeld \cite{klette}. Choose any $z\in\Z^n$ such that 
  $\beta(z)$ has a double point $p$. There is a (discrete) line
  $g(z,p)=\{q\in\Z^n:\tau^i(z)=q, i\in \Z\}$ with generator $\tau$
  through $z$ and $p$ with $\tau(p)=z$. Let $p'=\tau(z)$. So $p,z,p'$
  are three consecutive points of this line. Because of the regularity
  of $\beta$ the point $p'$ lies in $\beta(z)$.
 
  Furthermore $p'$ is a double point. In the case of the rotational
  invariance of $\beta$ this is immediately clear by rotation of $p$ in
  $\beta(z)$ and in the case of translational invariance the
  proposition follows by exchanging of the roles of $\tau(z)$ and
  $\tau(p)$ and the translation of the points $q$ and $r$ to $\tau(q)$
  and $\tau(r)$ in the definition \ref{doppelpunkt} of double point.

  The set $M:=\beta(z)\cup\{z\}$ is $N$-simply connected and can be
  reduced to the single point $p$ by repeated deletion of (simple)
  points. Therefore its Euler-characteristic is 
  \begin{equation}
    \chi(M)=1
  \end{equation}

  The set $M\SETMINUS\{z,p,p'\}$ is topologically equivalent to a
  circle. Thereby 
  \begin{equation}
   \chi(M\SETMINUS\{z,p,p'\})=0
  \end{equation}

  Let $E$ be an embedding of $\Z^2$ in $\Z^n$, generated by $\tau$ and
  the translation $\sigma$ between $q$ and $r$. $E$ contains the
  points $p,z,p'$. The set $M_0 := (M\SETMINUS\{p,z,p'\})\cap E$ is
  simple connected and after deletion of simple points a single point,
  such that $\chi(M_0)=1$.
  Let the sets $M^+$ and $M^-$ be the intersections of $M\SETMINUS\{p,z,p'\}$
  with the positive respectively negative closed (discrete) half-space
  of $E$. By symmetry holds $\chi(M^+)=\chi(M^-)=a$ and by the
  sum-formula of the Euler-characteristic is 
  \begin{equation}
    \chi(M\SETMINUS\{z,p,p'\})=\chi(M^+)+\chi(M^-)-\chi(M^+\cap
    M^-)=2a-1\neq 0
    \enspace .
  \end{equation}\qed
\end{Beweis}

\begin{Def}\label{goodpair}
  A separating pair of adjacencies $(\alpha,\beta)$ in
  $\Z^n$ is a \DF{good pair}, if for every $p\in\Z^n$ the set
  $\beta(p)$ contains no double points. 
\end{Def}


\section{Theorem of Jordan-Brouwer}

We want to prove a discrete variant of the theorem of
Jordan-Brouwer. But we haven't seen anything of it yet. This will
change right now.

\begin{Satz}[The Discrete Theorem of Jordan-Brouwer]
  Let the set $M\SUBSET\Z^n$ be a $(n-1)$-dimensional manifold under the pair $(\alpha,\beta)$ 
  with $n\ge 2$. Then $\Z^n\SETMINUS M$ has exactly two path-connected components
  and $S$ is their common boundary.\qed
\end{Satz}

This formulation is slightly more general than the original one, which only
deals with spheres. It follows immediately from  number 3.42 of chapter
XV of Alexandrovs book \cite{alexandrov}, which reads as

\begin{Satz}[Alexandrovs Theorem]
  Any $(n-1)$-dimen\-sional pseudo\-mani\-fold in $\R^n$ is
  orient\-able, dis\-connects $\R^n$ in exactly two path-com\-po\-nents and is
  the com\-mon boundary of both of them.\qed
\end{Satz}

We will see, how to construct a $(n-1)$-pseudomanifold in $\R^n$ from a
discrete $(n-1)$-manifold, thereby proving:

\begin{Satz}[Main Theorem]
  For every good pair $(\alpha,\beta)$ in $\Z^n$ with $n\ge 2$. The
  theorem of Jordan-Brouwer is true.
\end{Satz}

\subsection{The construction of a Simplicial Complex}

We start with the following construction:

Given an adjacency $\alpha$ and arbitrary points $p,q\in\Z^n$, we define:
\begin{equation}
  (p,q)_\alpha:=
  \left\{
  \begin{array}{cc}
    \{x\in\R^n:x=\lambda p+(1-\lambda)q,\lambda\in(0,1)\} &
    p\in\alpha(q)\\
    \emptyset & p\not\in\alpha(q)
  \end{array}
  \right.
\end{equation}
and
\begin{equation}
  [p,q] :=
  \{x\in\R^n:x=\lambda p+(1-\lambda)q,\lambda\in[0,1]\} 
  \enspace .
\end{equation}
These are the line segments in $\R^n$ between the points $p$ and 
$q$. The first equation resembles an edge of the adjacency $\alpha$ in
a geometric way for any $\alpha$.  

At first we construct the complex inside an arbitrary $n$-cube
$C^n$. Let $M\SUBSET C^n\SUBSET\Z^n$ be a $\alpha$-connected 
set, and $(\alpha,\beta)$ be a good pair. Then we define a simplicial
complex $K_{(\alpha,\beta)}(M)=K(M)$ by the following process:

For any $k$-cube $C=C^k$ with vertices $c_1,c_2,\ldots,c_{2^k}$
let 
\begin{equation}
\hat{c}=\sum_{i=1}^{2^k}\frac{c_i}{2^k}
\end{equation}
be the \DF{barycenter} of $\C$. Observe, that the barycenter need not
to lie in the discrete space containing the cube $\C$. But since every
such space can be embedded in a natural way in some Euclidean space,
the barycenter is defined.

Given some cube $C\SUBSET \Z^n$ of dimension $1\le k\le n$, we
evaluate the test $T(C)$ as true, if and only if 
\begin{equation}
\exists{p,q\in C}:(\hat{c}\in(p,q)_\alpha\land p,q\in M)\lor
(\hat{c}\not\in(p,q)_\beta\land \hat{c}\in[p,q]\land p,q\in\COMP{M})
\enspace.
\end{equation}
The test is true, if $C$ is a subset of $M$, if the
line segment between two $\alpha$-neighbors in $M\cap\C$ meet the barycenter of $C$,
or if the line segment between two points of $\COMP{M}\cap\C$, that are not
$\beta$-connected meets the barycenter of $C$. 

One could imagine the following situation: 
Two points $p,q\in M$ define a barycenter of some $k$-cube $C$ and at the same time two points
$p',q'\in \COMP{M}$ are $\beta$-connected and the segment $[p',q']$ meets this barycenter. 
This is a case we do not want but we will see in Lemma \ref{glatt-diskret}, 
that the concept of double points is the key to avoid this. 

We construct the complex $K(M)$ inductively. First, consider the
vertices:
\begin{equation}
  K^0(M) := M \cup\{\hat{c}: C\SUBSET M \mbox{ and $T(C)$ is true} \}
  \enspace.
\end{equation}

If the simplices of dimension $i-1$ are defined, we are able to
construct the $i$-simplices for all $C\SUBSET C^n$. We thereby consider only the
simplexes that are contained in the convex hull $\overline{C}\SUBSET\R^n$ of the 
cube $C\SUBSET\Z^n$:
\begin{equation}
  K^1(M) := \{ (x,\hat{c}):
  x \in K^{i-1}(M)\cap\overline{C}, \hat{c}\neq x\mbox{, $T(C)$ true }\}
  \enspace.
\end{equation}
\begin{equation}
  K^i(M) := \{(x_0,\ldots,x_{i-1},\hat{c}):
  (x_0,\ldots,x_{i-1})=\sigma\in K^{i-1}(M), \sigma\SUBSET\overline{C} \mbox{, $T(C)$ true }\}
  \enspace.
\end{equation}

This means, we connect the barycenters of the subcubes $C$ of $C^n$ that satisfy the 
test $T(C)$ to the already constructed $(i-1)$-simplices, thus forming $i$-simplices. Remember, 
that if the test $T(C)$ is true, then the barycenter of $C$ is a vertex of our complex $K(M)$.
 
The union of all $K^i(M)$ will be called $K(M)$.

\begin{Lemma}\label{kubus_komplex}
  The set $K(M)$ is a simplicial complex for every fixed pair $(\alpha,\beta)$
  and a set $M\SUBSET C\SUBSET\Z^n$.
\end{Lemma}

\begin{Beweis}
  We have to check the definition
  \ref{simplizialkomplex} of a simplicial complex. 

First, we show that every face $\sigma'$ of a simplex $\sigma=(x_0,\ldots,x_i)$ in
  $K(M)$ is an element of $K(M)$. Observe, that $\sigma$ is in
  $K^i(M)$ for some $i$. If $i=0$, so $\sigma$ has no proper faces and
  the proposition is true.

  Consider $i>0$. For every $j\in\{0,\ldots,i\}$ let
  \begin{equation}
    \sigma_j=(x_0,\ldots,x_{j-1},x_{j+1},\ldots,x_i)
  \end{equation}
  be the simplex formed by deletion of the $j$'th vertex.

  For $j=i$ the simplex $\sigma_j$ is by construction of $K^i$ 
  a member of $K^{i-1}$ and also lies in $K$.

  For $j<i$, $(x_0,\ldots,x_{j-1})$ is a simplex in $K^{j-1}$ and a face
  of $\sigma_j$. By construction the simplex $(x_0,\ldots,x_{j-1},x_{j+1})$ is 
  in $K^j$. Using induction, we see that the simplex
  $(x_0,\ldots,x_{j-1},x_{j+1},\ldots,x_i)$ is in $K^{i-1}$.

  Second, given different simplices $\sigma,\tau\in K(M)$ of the form
  $\sigma=(x_0,\ldots,x_k)$ and $\tau=(y_0,\ldots,y_j)$, we need to
  show their disjointness. Suppose the vertex sets
  $\{x_0,\ldots,x_k\}$ and $\{y_0,\ldots,y_j\}$ are disjoint. Then,
  $\sigma\cap\tau$ has to be empty, too. We prove this by induction over $k$. 

  If $k=0$, so the proposition is true. So let $k>0$ and let the claim
  be true for all proper subsets of $\{x_0,\ldots,x_k\}$. $\sigma$ arises
  from a $(k-1)$-simplex by attachment of a barycenter $x$ of a subcube of
  $C^n$. By assumption, $x$ is no vertex of $\tau$. By the inductive
  hypothesis is $(x_{i_0},\ldots,x_{i_{k-1}})\cap\tau=\emptyset$ for
  $i_0,\ldots, i_{k-1}\in\{0,\ldots,k\}$. This is the boundary of
  $\sigma$. Therefore, $\sigma\cap\tau=\emptyset$ cannot be true.

  If $\{x_0,\ldots,x_k\}\cap\{y_0,\ldots,y_k\}=\{p_0,\ldots,p_i\}$
  holds, then $\varrho=(p_0,\ldots,p_i)$ is a common proper face of
  $\sigma$ and $\tau$.

  None of the vertices of $\tau$ lies in the inside of $\sigma$ and
  vice versa, since the simplices arise by successively attaching
  barycenters to a vertex of $\C^n$. No face of $\tau$ can meet the
  inside of $\sigma$, because of the construction of the simplices.
\end{Beweis}

Let $M$ be a finite $\alpha$-connected subset of $\Z^n$ under the pair
$(\alpha,\beta)$. $M$ can be covered by a finite number of $n$-cubes
$C^n_1,\ldots,C^n_l$. We define:
\begin{equation}
  K(M):=\bigcup_{i=1}^lK(C_i^n\cap M)
  \enspace .
\end{equation}

\begin{Lemma}
  $K(M)$ is a simplicial complex for any finite set $M\SUBSET\Z^n$
  under the pair $(\alpha,\beta)$. 
\end{Lemma}

\begin{Beweis}
  The result follows immediately from the preceding lemma.
\end{Beweis}

If $M$ is not $\alpha$-connected but finite then we can construct
$K(M)$ component wise.

The next lemma is crucial in our theory, since it establishes the
connection between the topology of the discrete set $M$ and the
topology of the geometric realization $|K(M)|$ in $\R^n$.

\begin{Lemma}\label{glatt-diskret}
  Let $M\SUBSET\Z^n$ be a set under the pair
  $(\alpha,\beta)$ and $K(M)$ the simplicial complex from the
  construction above. Then the following properties hold:
  \begin{enumerate}  
  \item 
    \begin{equation}
      M=\Z^n\cap|K(M)|\enspace,
    \end{equation}
    \begin{equation}
      \COMP{M}=\Z^n\cap\COMP{|K(M)|}
      \enspace .
    \end{equation}
  \item Two points $p,q\in M$ are in the same $\alpha$-component of
    $M$ iff they are in the same component of $|K(M)|$. 
  \item Two points $p,q\in\COMP{M}$ are in the same $\beta$-component
    of $\COMP{M}$ iff they are in the same component of
    $\COMP{|K(M)|}$ and $(\alpha,\beta)$ does not contain double points.
  \item The boundary of a component $A\SUBSET|K(M)|$ meets the
    boundary of a component $B\SUBSET\COMP{|K(M)|}$ iff there exists a
    point in $A\cap\Z^n$ that is $\beta$-adjacent to some point in
    $B\cap\Z^n$.
  \end{enumerate}
\end{Lemma}

\begin{Beweis}
  1. To prove the first equation, we observe that a point $p$ is in $M$,
  iff it is a vertex of $K(M)$ and also a point of $\Z^n$. This is the
  case, iff $p$ is in $|K(M)|\cap\Z^n$. The second equation follows by
  an analog argument. 

  2. ($\IF$) Let $p,q$ be points from an $\alpha$-component of
  $M$. There exists an $\alpha$-path $p=p_0,p_1,\ldots,p_k=q$ such that
  $[p_i,p_{i+1}]_\alpha$ is in $|K(M)|$ for all $i\in\{1,\ldots,k-1\}$.
  This is a piecewise linear path.
  
  ($\ONLYIF$) Let $p,q$ be two points of the same component of
  $|K(M)|$. There exists a real path $w:[0,1]\rightarrow|K(M)|$ such that 
  $w(0)=p$ and $w(1)=q$. Let $\sigma_1,\ldots,\sigma_k$ be the
  simplices met by $w$. Every point of the path $w$ in a simplex
  $\sigma_i$ can be mapped to the boundary
  $\overline\sigma_i\SETMINUS\sigma_i$ by a linear homotopy.
  After a finite number of such homotopies the path $w$ is mapped to a
  path, which only meets edges and vertices of $K(M)$.
  This path can the be transformed into an $\alpha$-path by the points
  on it, that are in $\Z^n$.

  3. The proof is analog to 2. But the two directions do not hold in
  general if the pair has double points. Let $p,q,s,t$ be as in
  definition \ref{doppelpunkt} such that $p,q\in M$ and
  $s,t\in\COMP{M}$ are in different components of $\COMP{|K(M)|}$,
  $p\in\alpha(q)$ and $s\in\beta(t)$. The point
  $x=[p,q]_\alpha\cap[s,t]_\beta$ is in $K(M)$ by construction and
  therefore $x$ is in $|K(M)|$, too. 


  4. Let $A\SUBSET|K(M)|$ and $B\SUBSET\COMP{|K(M)|}$ be two components.
  
  ($\IF$) Suppose the intersection of the boundaries of $A$ and $B$ contains
  a point $x$. This point $x$ is in a $k$-simplex of $K(M)$ that has a vertex in
  a $n$-cube $C^n$, which meets $A$ and $B$. There are points $p,q\in
  C^n\cap\Z^n$, such that $p\in A\cap\Z^n\SUBSET M$ and $q\in
  B\cap\Z^n\SUBSET\COMP{M}$. In $C^n$ exists a $\pi$-path from $p$ to
  $q$. Since $\pi\SUBSET\beta$ this path need to contain two
  $\beta$-adjacent points such that one of them is in $A\cap\Z^n$ and the other is 
  in $B\cap\Z^n$.

  ($\ONLYIF$) Suppose there exist a $\beta$-path from $p\in A\cap\Z^n$ to
  $q\in B\cap\Z^n$. This path can be extended to a $\pi$-path by
  insertion of certain points from $\Z^n$. There exist two
  $\pi$-neighbors $p'\in A$ and $q'\in  B$ in $\Z^n$. By construction
  of $K(M)$ the point $p'$ has to be in the simplicial boundary of
  $K(M)$ and therefore in the intersection of the boundaries of $A$
  and $B$.
\end{Beweis}

\subsection{The Construction of a Pseudomanifold}

We now apply the construction of $K(M)$ to digital $(n-1)$-manifolds
$M$ with the goal to define a combinatorial $(n-1)$-pseudomanifold
(see definition \ref{pseudomannigfaltigkeit}). 

Let $M$ be a digital $(n-1)$-manifold in the rest of the section.
The first issue to be solved, is that $K(M)$ needs not to be homogenous
$(n-1)$-dimensional by construction, since $n$-simplices are easily
introduced.

\begin{figure}[htb]
\begin{center}
\includegraphics{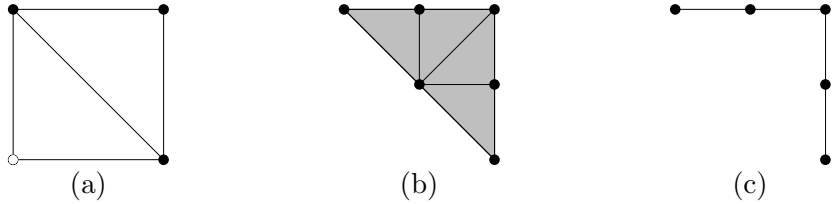}
\end{center}
\caption[Complex with $n$-simplex]{\small{How the complex $K'(M)$ is built.
    Figure (a) shows a 2-cube $C$ with corresponding adjacencies. The black
    dots comprise the set $M$. The points are used in figure (b) to build
    the complex $K(M)$. Since this complex may contain 2-simplices (depicted in gray),
    we need to find a strong deformation retract $K'(M)$ as in
    figure (c).}} 
\label{Pic:komplex_n_simplex}
\end{figure}

We solve this issue by finding a deformation retract $|K'(M)|$ of
$|K(M)|$, that has no $n$-simplices in $K'(M)$. The new complex $K'(M)$ is constructed as 
follows: At first we remove all vertices from $K(M)$ that are barycenters of the cubes $C$ 
with only one $\beta$-component in $C\SETMINUS M$. Then we remove
all simplices, that contain such a barycenter as vertex. 

The next lemma shows that this reduction preserves all the important topological properties of 
the initial complex $K(M)$.

\begin{Lemma}\label{sdr}
  The complex $K'(M)$ is a triangulation of a strong deformation retract of
  $K(M)$.
\end{Lemma}

\begin{Beweis}
  We begin by proving the next lemma:

  \begin{Lemma}
    Let $\sigma$ be a $k$-simplex and $x$ be any vertex of $K(\sigma)$.
    Then $|K(\sigma\SETMINUS\{x\})|$ is a strong deformation retract of
    $|K(\sigma)|$.
  \end{Lemma}
  
  \begin{Beweis}
    We give a linear homotopy from $|K(\sigma)|$ to
    $|K(\sigma\SETMINUS\{x\})|$. 

    Given $y\in|K(\sigma)|\SETMINUS\{x\}$ let $z$ be the point on the ray 
    $xy\SUBSET\R^n$ that lies in $\partial|K(\sigma)|$, that is the boundary 
    of $\sigma$. The function
    \begin{equation}
      f(y,t)=tz+(1-t)y
    \end{equation}
    is a linear homotopy between the given sets.
        
    $K(\sigma\SETMINUS\{x\})$ is a simplicial complex and by existence
    of $f$, the set $|K(\sigma\SETMINUS\{x\})|$ is a strong deformation retract
    of $|K(\sigma)|$. 
  \end{Beweis}
  
  Let now $f_\sigma$ be the homotopy defined above for every $\sigma\in K(M)$ 
  containing $x$ as vertex. The removal of $x$ from $\sigma$ induces 
  the following linear homotopy 

  \begin{equation}
    f(y,t)=\left\{\begin{array}{ccl}
    f_\sigma(y,t) & &\mbox{There exists a $\sigma$ with vertex $x$ containing $y$} \\
    y & & \mbox{otherwise}
    \end{array}\right.
    \enspace.
  \end{equation}

  Since only a finite number of points are removed from $M$, we can 
  consider the construction of $K'(M)$ a finite sequence of linear 
  homotopies of the form given above. Therefore, it follows that $K'(M)$
  is a strong deformation retract of the complex $K(M)$.
\end{Beweis}

\subsection{Properties of a Digital Manifold}

\begin{Lemma}\label{zwei-komp}
  Let $M\SUBSET\Z^n$, $n\ge2$, be a digital $(n-1)$-manifold under the pair $(alpha,\beta)$.
  For every $n$-cube $C\SUBSET\omega(M)\SUBSET\Z^n$, $n\ge 2$ the set
  $C\SETMINUS M$ contains at most one $\beta$-component in $C_M$ and
  $D_M$, respectively.
\end{Lemma}

\begin{Beweis}
  Since $M$ is a digital $(n-1)$-manifold, for all $p\in M$ the set
  $\omega(p)\SETMINUS M$ has exactly two $\beta$-components $C_p$ and $D_p$ 
  (property \ref{twocomponents}).

  Assume for contradiction $C\SETMINUS M$ has two $\beta$-components
  $K_1$ and $K_2$ in $C_M$.
  For every $p\in C\cap M$ the components $K_1$ and $K_2$ are contained in $C_p$ 
  and each $\beta$-path between $K_1$ and $K_2$ passes outside of $C$. 
  By Lemma \ref{komponentenschranken} exist in $C\cap
  M$ at least $n$ points, since $C\SETMINUS M$ contains at least two 
  $\beta$-components. Call these points $p_1,\ldots,p_n$. 

  Let $C'$ be a $(n-2)$-subcube of $C$, such that $C'\cap M$ has a 
  maximal number of points. There exist two translations $\tau_1$ 
  and $\tau_2$ such that
  \begin{equation}
    C=C'\cup\tau_1(C')\cup\tau_2(C')\cup\tau_1\tau_2(C')
  \end{equation}

  It is not possible to have for every choice of $C'$:
  \begin{equation}\label{eq::sep_prop}
    \FORALL_{p\in C'}\left((\tau_1(p)\in\tau_1(C')\SETMINUS M \lor
    \tau_2(p)\in\tau_2(C')\SETMINUS M) \THEN
    \tau_1\tau_2(p)\in\tau_1\tau_2(C')\SETMINUS M\right)
  \end{equation}
  since otherwise $C\SETMINUS M$ would only contain one $\beta$-component, 
  but $K_1$ and $K_2$ are two components.
  The set $C\cap M$ is by property \ref{cubeconnection} of $M$ 
  $\alpha$-connected and therefore, we have a contradiction to the separation 
  property, because the formula \ref{eq::sep_prop} is the core requirement of 
  this property.
\end{Beweis}

\begin{Folg}\label{zwei-komp-folg}
  Let $M\SUBSET\Z^n$, $n\ge2$, be a digital $(n-1)$-manifold under the pair $(alpha,\beta)$.
  Any $k$-cube $C$, $k\in\{0,\ldots,n\}$, with $C\cap M\neq
  \emptyset$ has at most two $\beta$-com\-po\-nents in $C\SETMINUS M$.
\end{Folg}

\begin{Beweis}
  This follows from Lemma \ref{zwei-komp} and Lemma
  \ref{dimensionskonsistenz}. 
\end{Beweis}

\begin{Lemma}\label{cubesequence}
  Let $M\SUBSET\Z^n$, $n\ge 2$, a digital $(n-1)$-manifold under the pair
  $(\alpha,\beta)$. Consider the sequence $C^0,\ldots,C^n$ of cubes ordered 
  increasingly by inclusion. The number of $\beta$-components in
  $C^i\SETMINUS M$ decreases at most by one in $\C^{i+1}\SETMINUS M$
  for all $i\in\{0,\ldots,n-1\}$. 
\end{Lemma}

\begin{Beweis}
  The proposition holds for $n$-cubes containing increasing sequences ordered 
  by inclusion that share no point with $M$. Corollary \ref{zwei-komp-folg} 
  guarantees the existence of at most two $\beta$-components in $C^n\SETMINUS M$ of 
  any such sequence $C^0,\ldots,C^n$.

  Assume for contraction that in a sequence $C^0,\ldots,C^n$ exists a $(j+1)$-cube
  $C^{j+1}$ that contains two more $\beta$-components in $\COMP{M}$ 
  than $j$-cube $C^{j}$. We know, that a translation $\tau_1$ exists 
  with $C^{j+1}=C^j\cup\tau_1(C^j)$.
  Furthermore a $(j-1)$-subcube $C^{j-1}$ of $C^j$ exists such that
  \begin{equation}
    C^{j+1}=C^{j-1}\cup\tau_1(C^{j-1})\cup\tau_2(C^{j-1})\cup\tau_1\tau_2(C^{j-1})
  \end{equation}
  for some translation $\tau_2$.
  The points of the two $\beta$-components of $C^{j+1}\SETMINUS M$ are contained 
  in $\tau_1(C^{j-1})$ and $\tau_1\tau_2(C^{j-1})$, since $C^j$ must be 
  completely contained in $M$.
  We cannot find a point $p$ in a $\beta$-component of $\tau_1(C^{j-1})\SETMINUS M$  
  for every possible choice of $C^{j-1}$ such that $\tau_2(p)$ lies 
  in $\tau_1\tau_2(C^{j-1})\SETMINUS M$, because otherwise, there could not be two
  $\beta$-components in $C^{j+1}\SETMINUS M$. But this clearly contradicts
  the separation property of $M$ as one can see in figure \ref{Pic:trennungseigenschaft}. 
  Furthermore, the number of $\beta$-components of a $C^j\SETMINUS M$ in a sequence 
  as given above, cannot exceed the number of $\beta$-components of 
  $C^{j+1}\SETMINUS M$. In this case two $\beta$-components $C^j\SETMINUS M$ 
  would be contained in one $\beta$-component of $C^{j+1}\SETMINUS M$, which is impossible
  as Lemma \ref{zwei-komp} shows. No other possibilities exist and therefore, the Lemma is 
  proven. 
\end{Beweis}

\begin{Lemma}
  Let $M\SUBSET\Z^n$, $n\ge2$, be a digital $(n-1)$-manifold under the pair $(alpha,\beta)$. 
  For every $n$-cube $C$ holds
  \begin{equation}
    |C\SETMINUS M|\ge 1
  \end{equation}
\end{Lemma}

\begin{Beweis}
  The proposition is true for $C\cap M=\emptyset$.
  Otherwise, we assume for contradiction that $C\SUBSET M$. Let $p$ be any point of $C$.
  Then, the set $M\SETMINUS p$ is $\alpha$-connected, since every subcube of
  $C\SETMINUS\{p\}$ is $\pi$-connected.
  By property \ref{componentunity} of the $(n-1)$-manifold $M$ 
  there are $n$ $\pi$-neighbors of $p$ $\beta$-adjacent to $C_p$ and $D_p$. Call these points 
  $p_1,\ldots,p_n$. 

  Let $\tau_1,\ldots,\tau_n$ be those generators of $\Z^n$ that satisfy 
  $\tau_i(p)=p_i$ and let $C_1,\ldots,C_n$ be the $(n-1)$-cubes with 
  $C_i\cap C=\emptyset$ and $\tau_i(C_i)\SUBSET C$. One of these cubes $C_i$
  has to contain two $\beta$-components in $\COMP{M}$, since all points
  $p_i$ in $C$ must be $\beta$-adjacent to $C_p$ and $D_p$.
  In the case $n=2$ it is clear that no such $C_i$ can exist. 
  For greater $n$ let w.l.o.g. be $C_n$ this cube. We know that in $C_i$ exists 
  a point of $M$. This point is $\beta$-adjacent to $\tau_i(C_i)\SUBSET C$ and
  $\tau_i(C_i)\cup C_i$ is a $n$-cube. We construct an increasing sequence of 
  cubes ordered by inclusion:
  \begin{equation}
    C^0=\{p\},C^1=\{p,\tau_1(p)\},C^{i+1}=C^i\cup\tau_{i+1}(C^i), C^n=\tau_n^{-1}(C^{n-1})
  \end{equation}
  The cubes $C^0,\ldots,C^{n-1}$ are completely contained in $C\SUBSET M$, but the cube
  $C_n$ contains two $\beta$-components in $\COMP{M}$ by assumption. By Lemma \ref{cubesequence}
  this is impossible. Therefore, we conclude that no $n$-cube lies completely 
  in $M$.
\end{Beweis}

\begin{Lemma}\label{kuben-komponenten}
  Let $M\SUBSET\Z^n$, $n\ge 2$, be a $(n-1)$-manifold under the pair $(\alpha,\beta)$. 
  Then every for $k$-cube $C\SUBSET M$ with $0\le k< n$ the set 
 \begin{equation}
   \bigcap_{p\in C}\omega(p)\SETMINUS M
 \end{equation}
  has two $\beta$-components. These are contained in $C_p$ and $D_p$ respectively 
  for every $p\in C$.
\end{Lemma}

\begin{Beweis}
  We use induction on $k$.

  For $k=0$ the result follows from property 1 of the $(n-1)$-manifold $M$.

  In the case $k>0$ let $C=C'\cup\tau(C')\SUBSET M$ be a $k$-cube for some 
  suitable translation $\tau$. The cube $C$ is contained in $M$ and the 
  $(k-1)$-cubes $C'$ and $\tau(C')$ satisfy the induction hypothesis. 
  The sets 
  \begin{equation}
    \bigcap_{p\in C'}\omega(p)\SETMINUS M
  \end{equation}
  and 
  \begin{equation}\label{eq::eq::eq}
    \bigcap_{p\in \tau(C')}\omega(p)\SETMINUS M
  \end{equation}
  each contain two $\beta$-components.

  In $\tau(C')$ lie the $\pi$-neighbors $q=\tau(p)$ for all $p\in C'$. Those points $q$ 
  possess $\beta$-neighbors in $\omega(p)\SETMINUS M$ by property 2 of $M$.
  It follows that they also contain $\beta$-neighbors in $\bigcap_{p\in C'}\omega(p)\SETMINUS M$. 
  The same holds for equation \ref{eq::eq::eq}.
  Thus, the set
  \begin{equation}
    \bigcap_{p\in C'}\omega(p)\cap\bigcap_{p\in \tau(C')}\omega(p)
  \end{equation}
  contains $n$-cubes, because $C$ and $C'$ only differ by a single translation $\tau$.
  These $n$-cubes have points of $\COMP{M}$ that belong to
  $D_p$ or $C_p$, respectively. These points are $\beta$-connected in the 
  corresponding $\beta$-components of $\bigcap_{p\in C}\omega(p)\SETMINUS M$.

  Since no $n$-cube can be contained in a $(n-1)$-manifold, the proposition holds
  for $k$-cubes $C$ with $k\le n-1$. The set $\bigcap_{p\in C}\omega(p)\SETMINUS M$
  therefore has two $\beta$-components.
\end{Beweis}

\begin{Lemma}\label{n-simplizes}
  Given any $(n-1)$-manifold $M\SUBSET\Z^n$, $n\ge 2$ under the pair $(\alpha,\beta)$
  the complex $K'(M)$ contains no $n$-simplices.
\end{Lemma}

\begin{Beweis}
  None of the $n$-cubes $C^n$ with $C^n\cap M\neq\emptyset$, can be contained 
  completely in $M$. Thus, in every increasing sequence of cubes ordered by 
  inclusion $C^0,\ldots,C^n$ exists at least one cube $C^j$ that has only one
  $\beta$-component in $\COMP{M}$.
  By construction, this cube $C^j$ has no barycenter in $K'(M)$ and thus the
  cubes of the sequence define at most $n$ barycenters, but a $n$-simplex is
  defined by $n+1$ points. 
\end{Beweis}


\section{The Proof of the Theorem of Jordan-Brouwer for
  Digital Manifolds}  

Having constructed the simplicial complex $K'(M)$, which is
guaranteed to contain no $n$-simplices, all we have to do now, is to prove
the next theorem. In the formulation of the theorem we use good pairs $(\alpha,\beta)$
although the definition of good pair itself relies on $(n-1)$-manifolds. If one checks 
decently, he will notice, that in no argument during the proof of 
Theorem \ref{n-1-pseudomannigfaltigkeit}, we use the property of $(\alpha,\beta)$ to
be a separating pair. So, we could require $(\alpha,\beta)$ to be just a pair without
double points and the correctness of the proof won't be affected. Furthermore, no 
circularity is introduced.

\begin{Satz}\label{n-1-pseudomannigfaltigkeit}
  Let $M\SUBSET\Z^n$ be a digital $(n-1)$-manifold under the good pair
  $(\alpha,\beta)$. 
  Then $K'(M)$ is a $(n-1)$-pseudomanifold.
\end{Satz}

\textbf{Proof:} We have to check that $K'(M)$ has the properties of a
$(n-1)$-pseudo\-mani\-fold. The proof is split into three parts, as 
Lemma \ref{n-1-simplex}, Lemma \ref{n-2-simplex} and Lemma \ref{strong-connection}. \qed


\subsection{The Homogeneity}

\begin{Lemma}[The Homogeneity]\label{n-1-simplex}
For every digital $(n-1)$-manifold $M\SUBSET\Z^n$ under
a good pair $(\alpha,\beta)$, $K'(M)$ is homogenous
$(n-1)$-dimensional. 
\end{Lemma}

\textbf{Idea of the proof:}
Given a simplex $\sigma$ we take a sequence of
cubes whose barycenters are the vertices of $\sigma$. If the dimension
of $\sigma$ is $n-1$, the lemma is true by reflexivity of the face realation. 
Otherwise, we know by the
properties of $M$ that in a sequence of $n+1$ cubes
$C^0,\ldots,C^n$, which is ordered by inclusion and meets $M$ in every
cube, exactly one cube looses its barycenter in the transition from
$K(M)$ to $K'(M)$. And so we can augment the sequence corresponding to
$\sigma$ by at least one cube with a barycenter in $K'(M)$. Thus,
defining inductively a $(n-1)$-simplex, that has $\sigma$ as a face.

\begin{Beweis}
  Let $\sigma$ be any $k$-simplex in $K'(M)$. We will prove the proposition
  by induction on $i=(n-1)-k$. Remember, that by Lemma \ref{n-simplizes} no $n$-simplices 
  can be contained in $K'(M)$.

  $i=0$. In this case $\sigma$ is a $n$-simplex and by reflexivity of the 
  face relation it is its own face. Therefore, the proposition is true.

  $i>0$. The simplex $\sigma$ is contained in a cube $C$ of minimal 
  dimension. Let the barycenter of $C$ be the last vertex that was inserted in $\sigma$
  during the construction of $K(M)$ and that was not deleted during the
  transition from $K(M)$ to $K'(M)$.

  We will show the existence of a cube $C^*$ whose barycenter $x$ lies in $K'(M)$ and is 
  no vertex of $\sigma$. The cube $C^*$ is either a sub- or supercube of $C$. 
  We can use $x$ to construct a $(k+1)$-simplex $\sigma'$ out of $\sigma$ such that 
  the induction hypothesis is satisfied. Since $\sigma$ is a face of $\sigma'$ by 
  transitivity of the face relation $\sigma$ is then contained in a $n$-simplex.

  Case 1: The simplex $\sigma$ shares no vertex with $C$. Then all vertices of
  $\sigma$ are barycenters of certain $l_i$-subcubes $C^{l_i}$ of $C$ with
  $0<l_i\le\dim(C)=k$ for $i=0,\ldots,k$. The vertices of $\sigma$ are
  contained in $K'(M)$, therefore the tests $T(C^{l_i})$ are true.
  Furthermore exists a $C^{l_i}$ with a vertex in $M$, since otherwise
  all of the $C^{l_i}$ are contained in $\COMP{M}$.
  Therefore a point $x\in M$ exists, which is the barycenter of a $0$-cube $C^*$ and we 
  may use $x$ to augment $\sigma$ to $\sigma'$.

  Case 2: The simplex $\sigma$ shares a vertex with $C$.

  Case 2.1: The dimension $\dim(C)=k'$ is greater than $k$. Remember, that by choice of 
  $C$ its barycenter $x$ is in $K'(M)$ and is also a vertex of $\sigma$.

  Case 2.1.1: If $C$ is a subset of $M$ then exists an increasing sequence of cubes ordered by
  inclusion, $C^0,\ldots,C^{k'}=C$, such that the barycenters of the cubes
  are the vertices of $\sigma$. Because of $k'\ge k+1$, in the sequence need to exist two
  consecutive cubes $C^i$ and $C^j$ with $j>i+1$. A $(i+1)$-cube $C^*$ that 
  contains $C^i$ and is contained in $C^j$ can be found. Since $C\SUBSET M$ all of 
  the cubes in the sequence are contained in $M$ and also $C^*$ is contained in $M$.
  By construction, the barycenter $x$ of $C^*$ ends up in $K'(M)$ and we may 
  construct $\sigma$ from $\sigma$.

  Case 2.1.2: If $C\SETMINUS M$ contains two $\beta$-components, not every cube of 
  the sequence $F=(C^0,\ldots,C^{k'}=C)$ is contained in $M$. By 
  Lemma \ref{cubesequence} $F$ splits into two subsequences.
  Let 
  \begin{equation}\label{eq::first_subsequence}
    C^0,\ldots,C^m
  \end{equation}
  the subsequence of all cubes of $F$ that are contained in $M$ and let
  \begin{equation}\label{eq::second_subsequence}
    C^{m'},\ldots,C^{k'}
  \end{equation}
  be the subsequence of cubes in $F$ that contain two $\beta$-components in 
  $\COMP{M}$.
  By Lemma \ref{dimensionskonsistenz} the relation $m<m'$ holds and the 
  dimensions of the cubes in the first sequence are at most $m$ and the dimensions
  of those in the second sequence are at least $m'$.

  Case 2.1.2.1: If in one of the two subsequences exist two consecutive cubes $C^i$ 
  and $C^j$ with $j<i+1$ then we can find a cube $C^*$ with $C^i\SUBSET C^*\SUBSET C^j$ 
  that has the same number of $\beta$-components in $\COMP{M}$ as $C^i$. Therefore, the 
  barycenter $x$ of $C^*$ ends up in $K'(M)$ and we may construct 
  $\sigma'$ from $\sigma$.

  Case 2.1.2.2: Suppose $k'<n$ then there exists a $(k'+1)$-cube $C^*$ that
  contains $C^{k'}$ and that has two $\beta$-components in $\COMP{M}$ by
  Lemma \ref{dimensionskonsistenz} and Lemma \ref{zwei-komp}. Its barycenter
  $x$ is in $K'(M)$ and we may construct $\sigma'$ from $\sigma$.

  Case 2.1.2.3: Otherwise $k'=n$. We are looking for a cube $C^*$ that contains $C^m$ and is 
  contained in $C^{m'}$ and has either none or two $\beta$-components in $\COMP{M}$.
  Assume for contradiction that no such $C^*$ exists. In this situation any 
  sequence
  \begin{equation}\label{eq::impossible_sequence}
    C^{m+1},\ldots,C^{m'-1}
  \end{equation}
  between $C^m$ and $C^{m'}$ has the property that all its cubes only have one
  $\beta$-component in $\COMP{M}$. 

  Since $\dim(\sigma)< n-1$ and $C^0,\ldots,C^n$ is the sequence of cubes
  whose barycenters are the vertices of $\sigma$, the values $m+1$ and $m'-1$ have to 
  be distinct. Therefore any sequence of the form \ref{eq::impossible_sequence} has at 
  least two Elements.

  Consider the following sets
  \begin{equation}
    S^{m+1}:=\{C^{m+1}:
    C^{m+1}=C^m\cup\tau(C^m)\land\mbox{$\tau$ is no generator of
      $C^{m}$}\} 
  \end{equation}
  \begin{equation}
    S^{m+2}:=\left\{C^{m+2}:\begin{array}{c}
      C^{m+2}=C^{m}\cup\tau_1(C^{m})\cup\tau_2(C^m)\cup\tau_1\tau_2(C_m) \\
      \mbox{and $\tau_1\neq\tau_2$ are no generators of
        $C^{m+1}$} \end{array}\right\}
  \end{equation}
  The set $S^{m+1}$ consists of all $(m+1)$-cubes that contain $C^m$ and 
  $S^{m+2}$ represents the set of $(m+2)$-cubes that contain $C^m$.
  We choose a $C'\in S^{m+1}$. Then, $C'$ is of the form $C^m\cup\tau(C^m)$ and the $m$-cube
  $\tau(C^m)$ contains at least one point of $\COMP{M}$, since $C^m\SUBSET M$ 
  and $C'$ has exactly one $\beta$-component in $\COMP{M}$ by assumption.

  The cubes $C''$ in $S^{m+2}$ possess two subcubes in $S^{m+1}$ by construction. These 
  subcubes are $C^m\cup\tau_1(C^m)$ and $C^m\cup\tau_2(C^m)$. The cube $\tau_1(C^m)$ 
  as well as the cube $\tau_2(C^m)$ contains at least one point in $\COMP{M}$ that belong to
  a $\beta$-component in $C''\SETMINUS M$. Since $M$ has the separation property and 
  $C^m\SUBSET M$, the $m$-subcube $\tau_1\tau_2(C^m)$ of $C''$ satisfies:
  \begin{equation}
    (\tau_1\tau_2)^{-1}(\tau_1\tau_2((C^m)\cap M)\SUBSET
    (\tau_1^{-1}(\tau_1(C^m)\cap M))\cap(\tau_2^{-1}(\tau_2(C^m)\cap
    M))
  \end{equation}
  Analog relations hold for $\tau_1\tau_3(C^m)$ and $\tau_2\tau_3(C^m)$. 
  Therefore, $C''\SETMINUS M$ contains only one $\beta$-component for each $C''$.

  Let $C'''$ any $(m+3)$-cube containing $C^m$. $C'''$ cannot contain
  two $\beta$-components in $\COMP{M}$, since $C'''$ has the following form
  for pairwise distinct simple translations $\tau_1,\tau_2,\tau_3$ that are no generators
  of $C^m$:
  \begin{equation}
    \begin{split}
      C''' = C^m &\cup\tau_1(C^m)\cup\tau_2(C^m)\cup\tau_3(C^m) \cup \\
           & 
             \cup\,\tau_1\tau_2(C^m)\cup\tau_1\tau_3(C^m)\cup\tau_2\tau_3(C^m)
             \cup \\
           &  \cup\,\tau_1\tau_2\tau_3(C^m).
    \end{split}
  \end{equation}
  by separation property the following holds
  \begin{equation}
    \begin{split}
  (\tau_1\tau_2\tau_3)^{-1}&(\tau_1\tau_2\tau_3((C^m)\cap M)\SUBSET \\
  &((\tau_1\tau_2)^{-1}(\tau_1\tau_2(C^m)\cap
  M))\cap((\tau_2\tau_3)^{-1}(\tau_2\tau_3(C^m)\cap M))
    \end{split}
  \end{equation}
  and 
  \begin{equation}
    \begin{split}
  (\tau_1\tau_2\tau_3)^{-1}&(\tau_1\tau_2\tau_3((C^m)\cap M)\SUBSET \\
  &((\tau_1\tau_3)^{-1}(\tau_1\tau_3(C^m)\cap
  M))\cap((\tau_2\tau_3)^{-1}(\tau_2\tau_3(C^m)\cap M))
    \end{split}
  \end{equation}
  and
  \begin{equation}
    \begin{split}
  (\tau_1\tau_2\tau_3)^{-1}&(\tau_1\tau_2\tau_3((C^m)\cap M)\SUBSET \\
  &((\tau_1\tau_2)^{-1}(\tau_1\tau_2(C^m)\cap
  M))\cap((\tau_1\tau_3)^{-1}(\tau_1\tau_3(C^m)\cap M)).
    \end{split}
  \end{equation}

  Together these observations yield the proposition that one of the
  $m$-cubes $C^m\cup\tau_i(C^m)$, $i=1,2,3$ has a maximal number of points in $M$.
  W.l.o.g let this be the one with $i=1$. We have
  \begin{equation}
    \begin{split}
    (\tau_2\tau_3)^{-1}&(\tau_2\tau_3(C^m\cup\tau_1(C^m))\cap M) \\
      & = (\tau_2\tau_3)^{-1}(\tau_2\tau_3(C^m) \cap M) \cup
      (\tau_2\tau_3)^{-1}(\tau_2\tau_3\tau_1(C^m) \cap M) \\
      &\SUBSET \tau_2^{-1}(\tau_2(C^m)\cap M)\cap\tau_3^{-1}(\tau_3(C^m)\cap
      M) \\
      &\qquad \cup \tau_2^{-1}(\tau_2\tau_1(C^m)\cap
      M)\cap\tau_3^{-1}(\tau_3\tau_1(C^m)\cap M) \\
      &= \tau_2^{-1}(\tau_2(C^m\cup\tau_1(C^m))\cap M)\cap\tau_3^{-1}
      (\tau_3(C^m\cup\tau_1(C^m))\cap M) 
    \end{split}
  \end{equation}
  This means that $C'''\SETMINUS M$ has only one $\beta$-component if $M$ has the 
  separation property and all subcubes $C'$ and $C''$ of $C'''$ contain 
  only one $\beta$-component in $\COMP{M}$.
  It is easy to extrapolate the argument to even higher dimensional cubes. This leads 
  to a contradiction to the form of the sequences \ref{eq::first_subsequence} and 
  \ref{eq::second_subsequence}.
  We may conclude, that each sequence \ref{eq::impossible_sequence} needs to 
  contain some cube $C^*$ that has either none or two $\beta$-components in $\COMP{M}$ and the 
  barycenter $x$ of  $C^*$ may be used to define $\sigma'$ out of $\sigma$.

  Case 2.2: Let the simplex $\sigma$ be contained in a cube $C$ of dimension $k$. Since $k\neq n$ 
  by induction, we only need to rule the case, that all cubes containing $C$ possess exactly one
  $\beta$-component in $\COMP{M}$, for in the other cases the cube $C^*$ 
  obviously exists.

  In this case $C$ itself is contained in $M$, since all the $k+1$ barycenters of the
  sequence of $k+1$ cubes are in $K'(M)$. It holds that $k< n-1$ because $k=n-1$
  is the base case of the induction. 
  Analog to the considerations in case 2.1.2.2 it is not possible that all
  cubes of dimensions $k+1$, $k+2$, and so on, containing $C$ only have one
  $\beta$-component in $\COMP{M}$, because $M$ has the separation property.

  Therefore, we can find a cube $C^*$ containing $C$ with either none or two
  $\beta$-com\-po\-nents in $\COMP{M}$ and we are able to construct $\sigma'$ from
  $\sigma$.
\end{Beweis}


\subsection{The Nondegenerateness}

\begin{Lemma}[The Nondegenerateness]\label{n-2-simplex}
  Let $M\SUBSET\Z^n$ for $n\ge 2$ be a $(n-1)$-manifold under the good pair $(\alpha,\beta)$. 
  Then every $(n-2)$-simplex $\sigma\in K'(M)$ is the face of exactly two $(n-1)$-simplices
  $\sigma_1,\sigma_2\in K'(M)$. 
\end{Lemma}

\textbf{Idea of the proof:} Given a $(n-2)$-simplex $\sigma$ we have
to construct on the base of $M$ two $(n-1)$-simplices $\sigma_1$ and 
$\sigma_2$ in $K'(M)$, each containing $\sigma$ as a face. This is done 
again by considering the sequence of cubes ordered by inclusion, whose 
barycenters are the vertices of $\sigma$. If we augment this sequence 
to a maximal sequence with $n+1$ cubes, only one of the cubes has to 
loose its barycenter during the transition from $K(M)$ to $K'(M)$. We 
will show, that there are always two possible choices for cubes to be inserted
during the augmentation, and these do not loose their barycenters. Then 
we can use these barycenters to define $\sigma_1$ and
$\sigma_2$ from $\sigma$.

\begin{Beweis}Let $C$ be the cube of minimal dimension containing $\sigma$.

  Case 1: $k=n$ The cube $C\SETMINUS M$ contains two $\beta$-components in $\COMP{M}$.
  Let 
  \begin{equation} \label{eq::cube_sequence}
    C^{i_0},C^{i_1},\ldots,C^n
  \end{equation}
  be the sequence of cubes ordered by inclusion whose barycenters define $\sigma$.

  Case 1.1: There exists an index $j$, $0\le j\le n-1$, such that $i_i+2=i_{j+1}$ and the number of 
  $\beta$-components of $C^{i_j}\SETMINUS M$ and $C^{i_{j+1}}\SETMINUS M$ is different.
  Because of $\dim(\sigma)=n-2$ either $i_0>0$ or there exists another index $j'$ with 
  $i_{j'}+2=i_{j'+1}$. If $i_0=0$ then the number of $\beta$-components of the cubes 
  $C^{i_{j'}}\SETMINUS M$ and $C^{i_{j'+1}}\SETMINUS M$ at the position $j'$ cannot change 
  because of Lemma \ref{dimensionskonsistenz} and Lemma \ref{zwei-komp}. This implies the 
  existence of two $(i_{j'}+1)$-cubes $C$ and $C'$ included between $C^{i_{j'}}$ and 
  $C^{i_{j'}+1}$.
  The barycenters of $C$ and $C'$ are therefore in $K'(M)$ and we can define two 
  $(n-1)$-simplices containing $\sigma$.

  If $i_0>0$ then $i_0=1$ and the two points of $C^{i_0}$ need to be in $M$ and thus
  in $K'(M)$. These two points are the barycenters we may use to define the two
  simplices $\sigma_1$ and $\sigma_2$.

  Case 1.2: An index $j$ exists with $i_j+3=i_{j+1}$. Since $\dim(\sigma)=n-2$, we have 
  $i_0=0$ in sequence \ref{eq::cube_sequence}. By property 4 of $M$ holds:
  \begin{equation}
    C^{i_l}\SUBSET M,\mbox{ for } 0\le l\le j
  \end{equation}
  and
  \begin{equation}
    C^{i_l}\SETMINUS M \mbox{ has two $\beta$-components},\mbox{ for } j<l\le n
  \end{equation}

  There exist three translations $\tau_1,\tau_2,\tau_3$, by which
  the cube $C^{i_{j+1}}$ can be generated from $C^{i_j}$.
  We define the following cubes
  \[C_1 := C^{i_j}\cup\tau_1(C^{i_j}) \]
  \[C_2 := C^{i_j}\cup\tau_2(C^{i_j}) \]
  \[C_3 := C^{i_j}\cup\tau_3(C^{i_j}) \]
  and
  \[C_{12} := C^{i_j}\cup\tau_1(C^{i_j})\cup\tau_2(C^{i_j})
  \cup\tau_1\tau_2(C^{i_j}) \] 
  \[C_{13} := C^{i_j}\cup \tau_1(C^{i_j})\cup\tau_3(C^{i_j})
  \cup\tau_1\tau_3(C^{i_j}) \] 
  \[C_{23} := C^{i_j}\cup\tau_2(C^{i_j})\cup\tau_3(C^{i_j})
  \cup\tau_2\tau_3(C^{i_j}) \]
  and examine the relations between these cubes and M:

  Case 1.2.1: $C_1,C_2,C_3\SUBSET M$ or $\C_{12},C_{13},C_{23}$ each contain exactly one 
  $\beta$-component in $\COMP{M}$. The two situations are dual. This can not be the case:

  Because of the assumption on $C_{xy}$, $x,y=1,2,3$, $x<y$, we can conclude
  \begin{equation}
    \begin{array}{ccccccccc}
      C_{xy} & = &C^{i_j} & \cup & \tau_x(C^{i_j}) & \cup &
    \tau_y(C^{i_j}) & \cup & \tau_x\tau_y(C^{i_j})
    \\ \\
    & & \SUBSET M & & \SETMINUS M\neq\emptyset & & \SETMINUS
    M\neq\emptyset & & \cap M\neq\emptyset \\
    \end{array}
  \end{equation}
  Since $M$ has the separation property and $C^{i_j}\SUBSET M$ the relation
  \begin{equation}
    (\tau_x\tau_y)^{-1}(\tau_x\tau_y(C^{i_j})\cap M) \SUBSET
    \tau_x^{-1}(\tau_x(C^{i_j})\cap
    M)\cap\tau_y^{-1}(\tau_y(C^{i_j})\cap M)
  \end{equation}
  holds. Analog to the case 2.1.2.2 of the last proof $C^{i_{j+1}}\SETMINUS M$
  cannot contain two $\beta$-components. But this is a contradiction to the precondition 
  on $i_{j+1}$.

  Case 1.2.2: Let $C_1,C_2\SUBSET M$. By case 1.2.1 and Lemma \ref{cubesequence} the set
  $C_3\SETMINUS M$ has to contain one $\beta$-component. The cube $C_{12}$ must have
  only one $\beta$-component in $\COMP{M}$ by Lemma \ref{cubesequence}. It remains to
  show that the both cubes $C_{13}$ and $C_{23}$ each contain only one $\beta$-component in $\COMP{M}$.

  The subcube $\tau_3(C^{i_j})$ possesses points in $\COMP{M}$. We take a look at $C_{13}$.
  If this cube had two $\beta$-components in $\COMP{M}$ then its points would be contained in
  $\tau_1(C^{i_j})$ and $\tau_1\tau_3(C^{i_j})$, respectively, since $C_1\SUBSET M$.
  But then the sequence 
  \begin{equation}
    C^0,\ldots,C^{i_j},C_1,C_{13},C^{i_{j+1}},\ldots,C^n
  \end{equation}
  would contradict the Lemma \ref{cubesequence}, because the cubes $C_1$ and $C_{13}$
  differ by two $\beta$-com\-po\-nents. An analog argument can be used for $C_{23}$.
  Therefore, only the two barycenters of $C_1$ and $C_2$ exist in $K'(M)$ and thus we can define 
  the two simplices $\sigma_1$ and $\sigma_2$.

  Case 1.2.3: Let $C_1$ be a subset of $M$ and let the cubes $C_2,C_3$ contain only one 
  $\beta$-component in $\COMP{M}$. Then $C_{12}$ and $C_{13}$ both contain only one 
  $\beta$-component.
  The argument is analog to case 1.2.2. If $C_{23}$ contains only one $\beta$-component
  in $\COMP{M}$ then we could argument like in case 1.2.1 and build a contradiction to the
  fact that $C^{i_{j+1}}\SETMINUS M$ has two $\beta$-components.

  The barycenters of $C_1$ and $C_{23}$ are the points in question. We use them 
  to construct the $(n-1)$-simplices $\sigma_1$ and $\sigma_2$ from $\sigma$.

  Case 1.2.4: The cubes $C_{12}$ and $C_{13}$ contain two $\beta$-components in $\COMP{M}$.
  Because of case 1.2.1 and property 4 of $M$ the cube $C_{23}$  can only contain 
  one $\beta$-component in $\COMP{M}$. We have to show that $C_1,C_2,C_3$ each have only
  one $\beta$-component in $\COMP{M}$.

  The cube $\tau_1(C^{i_j})$ cannot be contained in $M$ since the sequence
  \begin{equation}
    C^0,\ldots,C^{i_j},C_1,C_{12},C^{i_{j+1}},\ldots,C^n
  \end{equation}
  would lead to a contradiction to Lemma \ref{dimensionskonsistenz}. An analog argument holds
  for $\tau_2(C_j)$ and $\tau_3(C_j)$, but by using $C_2$ and $C_{12}$ respectively
  $C_3$ and $C_{13}$.

  The cubes $C_1,C_2$ and $C_3$ can not contain two $\beta$-components in $\COMP{M}$ 
  since then we would reach a contradiction to Lemma \ref{dimensionskonsistenz}, for instance 
  with the sequence
  \begin{equation}
    C^0,\ldots,C^{i_j},C_1,C_{12},C^{i_{j+1}},\ldots,C^n
  \end{equation}
  We use the barycenters of $C_{12}$ and $C_{13}$ to construct the $(n-1)$-simplices
  $\sigma_1$ and $\sigma_2$.

  Case 1.2.5: Let $C_{12}\SETMINUS M$ contain two $\beta$-components and let the sets
  $C_{13}\SETMINUS M$ and $C_{23}\SETMINUS M$ have only one. Therefore, the cubes $C_1$ 
  and $C_2$ need to possess only one $\beta$-component in $\COMP{M}$, since they are subcubes
  of $C_{12}$. This means, that points from $\COMP{M}$ are contained in 
  $\tau_1(C^{i_j})$ and $\tau_2(C^{i_j})$. If $C_3\SUBSET M$ has only one
  $\beta$-component we would reach a contradiction as in case 1.2.1, since 
  $C_1,C_2$ and $C_3$ would then contain only one $\beta$-component in $\COMP{M}$.

  Therefore, $C_3$ is a subset of $M$, and the barycenters needed to construct $\sigma_1$ and
  $\sigma_2$ stem from $C_{12}$ and $C_3$.

  Case 2: $k<n$.

  Case 2.1: The cube $C$ lies in $M$ and $k=n-1$. Then, in the sequence
  \begin{equation}
    C^{i_0},C^{i_1}\ldots,C^{i_{k-1}},C^{n-1}
  \end{equation}
  holds either $i_0=0$ or $i_0=1$. In the case $i_0>0$ we can argument as in case 1.1.
  
  In the case $i_0=0$ there exists an index $j\in\{0,\ldots,k-1\}$ such that 
  $i_{j+1}=i_j+2$ since $\sigma$ has dimension $n-2$. There exist exactly two 
  $(i_j+1)$-cubes $C_1$ and $C_2$ that are contained in $M$, just like all the other
  cubes of the sequence. The barycenters of $C_1$ and $C_2$ let us define $\sigma_1$
  and $\sigma_2$.

  Case 2.2: The set $C\SETMINUS M$ has two $\beta$-components in $\COMP{M}$ and $k=n-1$.
  Then two cubes exist that contain $C$ and both of them need to have two $\beta$-components in 
  $\COMP{M}$. We can use their barycenters to define $\sigma_1$ and $\sigma_2$.

  Case 2.3: Let $C$ lie in $M$ and $k=n-2$. By Lemma \ref{kuben-komponenten}, there are 
  exactly two $\beta$-components in $\bigcap_{p\in C} \SETMINUS M$, that are contained 
  in $C_p$ and $D_p$ for all $p\in C$. The set $\bigcap_{p\in C} \SETMINUS M$ contains
  four $(n-1)$-cubes $C'_1,\ldots,C'_4$ and four $n$-cubes $C''_1,\ldots,C''_4$ that contain
  $C$. We can only have one of the following cases:
  \begin{enumerate}
  \item  There are two $(n-2)$-cubes $C_1$ and $C_2$ completely contained in $M$.
    If there were more, then $\bigcap_{p\in C} \SETMINUS M$ would contain more than
    two $\beta$-components. If there were just one such, then we had only one 
    $\beta$-component.
  \item There are exactly two $n$-cubes $C_1$ and $C_2$ containing two $\beta$-components 
    in $\COMP{M}$. With analog justification as in the case above.
  \item There is exactly one $(n-2)$-cube $C_1$ contained in $M$ and exactly one
    $n$-cube $C_2$ containing two $\beta$-components in $\COMP{M}$. Again
    with analog explanation as in the case above.
  \end{enumerate}
  in any of the three cases we can construct the two $(n-1)$-simplices $\sigma_1$ 
  and $\sigma_2$ from the cubes we found.
\end{Beweis}


\subsection{The strong connectivity}

\begin{Lemma}[The Strong Connectivity]\label{strong-connection}
  Let $M\SUBSET\Z^n$ for $n\ge 2$ be a $(n-1)$-manifold under the good pair 
  $(\alpha,\beta)$. Then $K'(M)$ is strongly connected.
\end{Lemma}

\textbf{Idea of the proof:} Given any two $(n-1)$-simplices
$\sigma$ and $\sigma'$, we know that each of them must have a vertex $p$
and $p'$, respectively, in the set $M\SUBSET\Z^n$. Since $M$ is
$\alpha$-connected, we can find a path $P$ from $p$ to $p'$, which
meets some $n$-cubes $C_0,\ldots,C_m$. We will show that
this sequence of $n$-cubes can be chosen such that two consecutive
cubes share a common $(n-1)$- or a common $(n-2)$-subcube. We are able
to construct sequences of $(n-1)$-simplices with the desired property,
first inside any of these cubes and even in between of them. As a rule
of thumb, one can say that the property of the simplices follows from
the same property of the cubes covering the $(m-1)$-manifold $M$.

We first prove the following lemma:

\begin{Lemma}\label{simplecase}
  Let $M\SUBSET\Z^n$ for $n\ge 2$ be a $(n-1)$-manifold and let $C_0,\ldots,C_l$ be
  a sequence of $k$-cubes, $1\le k\le n-1$, with $C_i\SUBSET M$ for $0\le i \le l$ 
  such that $C_i\cap C_{i+1}$ is a $(k-1)$-cube for all $i\in\{0,\ldots,l-1\}$.

  There exists a sequence $\sigma_0,\ldots,\sigma_{l'}$ of $k$-simplices in $K'(M)$
  with $\sigma_0\SUBSET C_0$ and $\sigma_{l'}\SUBSET C_l$ such that $\sigma_i$ 
  and $\sigma_{i+1}$ for all $i\in\{0,\ldots,l'-1\}$ share a common $(k-1)$-simplex.
\end{Lemma}

\begin{Beweis} We use induction on $k$.

  $k=1$. The sequence $C_0,\ldots,C_l$ describes a $\pi$-path in $M$ that induces a
  path in $K'(M)$. This path in $K'(M)$ is the sequence of 1-simplices we are looking 
  for. For an idea on the construction of this path take a look at figure \ref{Pic:kantenzug}.

  \begin{figure}[htb]
  \begin{center}
  \includegraphics{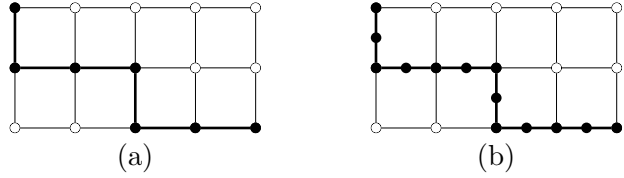}
  \end{center}
  \caption[$\pi$-path with corresponding complex]{\small{A $\pi$-path in
      figure (a) is transformed in a path in the corresponding simplicial complex
      in figure (b).}} 
  \label{Pic:kantenzug}
  \end{figure}

  $k>1$. Let $C_0,\ldots,C_l$ be a sequence of $k$-cubes satisfying the precondition of the Lemma.
  Let $\sigma_i$ be the simplex we found last in our already constructed sequence. In 
  $C_0$ such a simplex exists. This is $\sigma_0$ in the beginning. We need to find
  a sequence of $k$-simplices leading from $C_i$ to $C_{i+1}$. We can use the same argument 
  for all cubes of the sequence. Let $x_i$ be the barycenter of $C_i$

  The $(k-1)$-simplex $\sigma_i\SETMINUS x_i$ is a face of $\sigma_1$. By construction it 
  is contained in a $(k-1)$-subcube $C'_0$ of $C_i$. Let $C'_j$ be the $(k-1)$-cube 
  $C_i\cap C_{i+1}$. Obviously, In $C_i$ exists a sequence of subcubes $C'_0,\ldots,C'_j$ 
  satisfying the induction hypothesis. Therefore exists a sequence 
  $\sigma_i\SETMINUS x_i=\sigma_0',\ldots,\sigma_j'$ of $(k-1)$-simplices and 
  two consecutive elements of this sequence share a common $(k-2)$-face.

  By augmenting all $(k-1)$-simplices of this sequence by $x_i$ we can construct the
  sequence of $k$-simplices we are looking for. The simplex $\sigma_j'$ can also be augmented
  by $x_j$, the barycenter of $C_j$. The result lies in $C_j$ and we can iterate the construction
  with $(\sigma_j', x_j)$ in the role of $\sigma_i$.
\end{Beweis}

\begin{Lemma}\label{hardercase}
  Let $M\SUBSET\Z^n$ for $n\ge 2$ be a $(n-1)$-manifold and let
  $C_0,\ldots,C_l$ be a sequence of $k$- and $(k-1)$-cubes. Let the $k$-cubes of
  the sequence have two $\beta$-components in $\COMP{M}$ and let the $(k-1)$-cubes be 
  be each completely contained in $M$ for all $k$ with $2\le k\le n$. Suppose the sequence 
  satisfies:
  \begin{enumerate}
  \item The intersection of two consecutive $k$-cubes is either a $(k-2)$-cube
    contained in $M$ or a $(k-1)$-cube with two $\beta$-components in $\COMP{M}$.
  \item The intersection of two consecutive $(k-1)$-cubes is a $(k-2)$-cube contained 
    in $M$.
  \item The intersection of a $k$-cube and a subsequent $(k-1)$-cube
    is a $(k-2)$-cube contained in $M$ (and vice versa).
  \end{enumerate}

  Then exists a sequence $\sigma_0,\ldots,\sigma_{l'}$ of $(k-1)$-simplices such that
  $\sigma_0\SUBSET C_0$, $\sigma_{l'}\SUBSET C_l$ and $\sigma_i$ and $\sigma_{i+1}$ share a common
  $(k-2)$-face for $i\in\{0,\ldots,l-1\}$.
\end{Lemma}

\begin{Beweis} We use induction on $k$.

  $k=2$.
  \begin{enumerate}
  \item Subsequences of 2-cubes with two $\beta$-components in
    $\COMP{M}$ define an $\alpha$-path that induces a path in $K'(M)$. This path in
    $K'(M)$ is a subsequence of simplices we are looking for.
  \item Subsequences of 1-cubes can be treated as in Lemma \ref{simplecase}.
  \item We now consider the case of the transition between 2-cubes and 1-cubes.
    W.l.o.g let $C_i$ be a 2-cube with two $\beta$-components and let $C_{i+1}$ be
    the next $(k-1)$-cube of the sequence. It is contained in $M$ .
    $C_i\cap C_{i+1}$ is a single point $p$ in $M$. We get from the last 1-simplex 
    $\sigma$ in $C_i$ to the first 1-simplex in $C_{i+1}$ by exchanging the 
    barycenter of $C_i$ in $\sigma$ with the barycenter of $C_{i+1}$.
  \end{enumerate}

  $k>2$.
  \begin{enumerate}
  \item Let $C_i$ and $C_{i+1}$ be two consecutive $k$-cubes in the sequence. Each cube has 
    two $\beta$-components in $\COMP{M}$. Let $\sigma$ be the last $(k-1)$-simplex
    of the so far constructed sequence of simplices. We describe the transition from
    $C_i$ to $C_{i+1}$. Let $x_i$ and $x_{i+1}$ be the barycenters of $C_i$ and $C_{i+1}$, 
    respectively.

    Case 1: $C_i\cap C_{i+1}$ is a $(k-2)$-cube contained in $M$. Analog is the case that
    the intersection is a $(k-1)$-cube with exactly one $\beta$-component in $\COMP{M}$.

    $\sigma_i$ has a vertex $p$ in $M$ and the set $C_i\cap C_{i+1}$ contains a point $p'\in M$.
    There exists an $\alpha$-path from $p$ to $p'$ since $M$ is $\alpha$-connected. As one 
    can easily see, we may choose a path that lies in certain $(k-1)$-subcubes 
    $C_0',\ldots,C_j'$ of $C_i$ in this order, such that the Lemma \ref{simplecase} or 
    the induction hypothesis is satisfied. The cube $C_j'$ lies in $C_i\cap C_j$. 

    By induction hypothesis we have a sequence $\sigma_0',\ldots,\sigma_j'$ of 
    $(k-2)$-simplices such that consecutive elements share a common $(k-3)$-face. The 
    augmentation of the elements of the sequence by the barycenter $x_i$ of $C_i$ yields 
    a sequence of $(k-1)$-simplices with the desired property. This sequence lies 
    completely in $C_i$. 
    We have to do the transition into $C_{i+1}$ yet. This is simply done by exchanging the
    vertex $x_i$ in $\sigma_j'$ by the barycenter $x_{i+1}$ of $C_{i+1}$. Thus, the desired 
    subsequence of simplices is constructed.

    Case 2: The cube $C_i\cap C_{i+1}$ is a $(k-1)$-simplex with two $\beta$-components 
    in $\COMP{M}$. We proceed analogue to the case above with the only difference that 
    the cube $C_j$ in the sequence $C_0,\ldots,C_j$ is now a $(k-1)$-cube with two
    $\beta$-components in $\COMP{M}$.
  \item The case of a subsequence of $(k-1)$-cubes is handled by \ref{simplecase}.
  \item The transition from a $k$-cube $C_i$ to a $(k-1)$-cube $C_{i+1}$ can be
    handled like the first case. But the cube $c_j$ of the sequence $C_0',\ldots,C_j'$
    is a $(k-2)$-cube completely in $M$.
  \end{enumerate}
\end{Beweis}

We will now use the two lemmata to prove the strong connectivity of $K'(M)$.

\begin{Beweis}
  We will try to create a situation where Lemma \ref{hardercase} applies.

  Let $\sigma$ and $\sigma'$ be any two $(n-1)$-simplices in $K'(M)$. The simplices $\sigma$
  and $\sigma'$ each contain a vertex in $M$. Let these vertices be $p$ and $p'$. Since $M$
  is $\alpha$-connected, we can find an $\alpha$-path $P$ from $p$ to $p'$. The path $P$
  meets certain $n$-cubes $C_0\ldots,C_l$ in this order and these cubes either contain one or 
  two $\beta$-components in $\COMP{M}$.

  Let $r$ and $q$ be any two consecutive points on the path $P$. We show how to construct
  a sequence of cubes satisfying the preconditions of Lemma \ref{hardercase} locally
  with the help of $r$ and $q$. The method can be extended to the whole path $P$.

  The path $P$ induces a path in $K'(M)$ that contains 1-simplices $\sigma_1'$ and $\sigma_2'$. 
  These two simplices possess the vertices $r$ and $q$, respectively. By Lemma \ref{n-1-simplex} 
  two simplices $\sigma_1$ and $\sigma_2$ do exist, having the respective 1-simplex as a face.

  Case 1: The simplices $\sigma_1$ and $\sigma_2$ may be found in a 
  common $n$-cube $C$.

  Case 1.1: The set $C\SETMINUS M$ has two $\beta$-components. Therefore, the sequence comprised 
  of the single cube $C$ satisfies the premisses of Lemma \ref{hardercase}. Thus
  we can construct the sequence of simplices in question.

  Case 1.2: The set $C\SETMINUS M$ has only one $\beta$-component. Since $\sigma_1$ and $\sigma_2$
  are contained in $C$ and the barycenter of $C$ is not in $K'(M)$, the simplices have
  to be situated in some $(n-1)$-cubes $C_1,C_2\SUBSET M$.

  If $C_1=C_2$, then the $C_1$ is a sequence such that Lemma \ref{hardercase} may be applied.
  Otherwise $C_1$ and $C_2$ contain a common $(n-2)$ subcube. If this was not the case,
  $C_1$ and $C_2$ would differ by a translation and $C$ would be situated in $M$. But 
  this is impossible.
  Thus the afore-mentioned lemma may be applied.

  Case 2: The $n$-simplices $\sigma_1$ and $\sigma_2$ can only be located in different cubes.
  These cubes must have dimension $n$ or $n-1$. 

  Since $r$ and $q$ are consecutive points of $P$ they have to be $\alpha$-neighbors.
  This means that they are contained in a common $k$-cube $C$, $1<k<n$. 
  Let $k$ be minimal. Assume for contradiction that no other point of $M$ is in $C$.

  The case $k=1$ only happens for $n>2$, because for $n=2$ in $C$ obviously exists a
  $n-1=k$-cube containing the two 1-simplices. The precondition of case 2 disallows
  a $n$-cube with vertices $r$ and $q$ that contains the $n$-simplices $\sigma_1$ and 
  $\sigma_2$. Therefore, any $n$-cube and even any $(n-1)$-cube containing $C$ encloses 
  exactly one $\beta$-component in $\COMP{M}$. The set 
  $\omega(r)\cap\omega(q)\SETMINUS M$ inherits this property and thus only has one 
  $\beta$-component. This contradicts the property \ref{componentunity} of a 
  $(n-1)$-manifold. 

  In the case $k\ge 2$ for the $k$-cube $C$, the manifold $M$ is separated by $\COMP{M}$ 
  in $C$. To see this, we argue that $r$ and $q$ have to be located on diagonal vertices 
  in $C$, since $C$ has minimal dimension and no other points of $M$ are contained in $C$. 
  Let $C^*$ be any $(k-2)$-subcube of $C$ with $r\in C^*$ and let $\tau_1,\tau_2$ be the 
  translations that generate $C$ from $C^*$. Then $\tau_i(C^*)\cap M$ is empty 
  but $\tau_1\tau_2(C^*)$ contains $q$. This contradicts the separation property of $M$.

  We conclude that at least one additional point $s$ is in $C$ and this point is 
  $\alpha$-connected to $r$ and $q$. Therefore, we are able to augment the path $P$
  locally by such points $s$ between $r$ and $q$. We do not state here, that $s$ is 
  $\alpha$-adjacent to $r$ and $q$ but from property \ref{cubeconnection} of $M$ 
  follows the existence of a local path $r,s_1,\ldots,s_m,q$ for some $m\ge 1$.
  
  We now consider only one point $s$ with predecessor $r$ and successor $q$.
  The points $q,r$ are contained in $\omega(s)$ and so the simplices $\sigma_1$ 
  and $\sigma_2$ are located in two cubes $C'$ and $C''$, respectively, of dimension 
  $n$ or $n-1$. The cubes $C'$ and $C''$ are situated in $\omega(s)$. 
  Such a situation can always be achieved, since the $\alpha$-adjacency of 
  consecutive points induces 1-simplices that are faces of $n$-simplices by 
  Lemma \ref{n-1-simplex}. Also compare this to the first case of this proof.

  We now will construct a sequence of cubes $F = (C_0,\ldots, C_l)$ inside the set 
  $\omega(s)$. Let $C'$ be the last cube we explored. At the beginning holds $F=(C')$. 
  The cube $C''$ will always be the goal-cube.
  We also give every cube we already explored a mark.

  Case 2.1: $\dim(C')=\dim(C'')=n$. 

  If the set $C^*=C'\cap C''$ contains a $(n-1)$-cube
  with two $\beta$-components in $\COMP{M}$ or a $(n-2)$-cube in $M$, then the cube $C''$
  is appended to $F$. The construction of $F$ has come to an end and $F$ satisfies
  the premisses of Lemma \ref{hardercase}.

  Else, let $C$ be a $n$-cube of $\omega(s)$ that is not yet in $F$ and not yet marked. 
  Let $C^\#$ be the cube $C\cap C'$. If $C\SETMINUS M$ and $C^\#\SETMINUS M$ contain two 
  $\beta$-components, then append $C$ to $F$ and iterate the process with
  $C$ in the place of $C'$.

  Else, all the $n$-cubes $C$ of $\omega(s)$ that are not yet in $F$, that are not 
  marked and share a common $(n-1)$-cube with $C''$, have one $\beta$-component in $\COMP{M}$.

  The $n$-cube $C'$ then contains a $(n-2)$-subcube $C^\#\SUBSET M$. This holds, since $C'$ 
  contains a $(n-1)$-simplex $\sigma$ with vertex $s$ and $C^0,\ldots,C^\#,C^*,C'$ is a sequence of 
  cubes ordered by inclusion of the length $n+1$. The elements of the sequence are 
  corresponding with the vertices of $\sigma$ except for one (that is $C^*$). 
  The set $C'\SETMINUS M$ has two $\beta$-components and $C^*\SETMINUS M$ 
  has exactly one, otherwise we were in another case. 
  We conclude that $C^\#$ is contained in $M$. By Lemma \ref{kuben-komponenten}
  we can find two $\beta$-components in $\bigcap_{p\in C^\#}\omega(p)\SETMINUS M$
  and thus, in $\bigcap_{p\in C^\#}\omega(p)$ exists a cube $C$ with one of the properties
  \begin{equation}
    \dim(C)=n-1, C\SUBSET M
  \end{equation}
  or
  \begin{equation}
    \dim(C)=n, C\SETMINUS M\mbox{ has two $\beta$-components}\enspace.
  \end{equation}
  If $C$ is not already in $F$ and $C$ is not already marked, then append $C$ to $F$
  and iterate the process.

  Else, all the cubes $C$ of $\omega(s)$ are already in $F$ or marked. We look for the last 
  cube $D$ in $F$ such that one of the cases 2.1 to 2.4 is applicable. We remove all of the 
  cubes between $D$ and $C'$, mark them and let $D$ take the role of $C'$. We will show 
  in Lemma \ref{kubus-existenz} that such a cube $D$ always exists.

  Case 2.2: $\dim(C')=n-1$ and $\dim(C'')=n$. In this case holds $C'\SUBSET M$
  and $C''$ has one or two $\beta$-components in $\COMP{M}$.
  If $C^*=C'\cap C''$ is a $(n-2)$-cube in $M$ then 
  $F$ is completely constructed and Lemma \ref{hardercase} may be applied.

  Else we can find in $C'$ a $(n-2)$-cube $C^\#\SUBSET M$. Then the set $\bigcap_{p\in
    C^\#}\omega(p)\SETMINUS M$ has two $\beta$-components.

  If in the set $\bigcap_{p\in C^\# }\omega(p)$ exists a cube $C$ with
  \begin{equation}
    \dim(C)=n-1, C\SUBSET M
  \end{equation}
  or
  \begin{equation}
    \dim(C)=n, C\SETMINUS M\mbox{ has two $\beta$-components}
  \end{equation}
  that is not yet in $F$ and not yet marked, then append it to $F$,
  let $C$ play the role of $C'$ an iterate the process.

  Case 2.3: $\dim(C')=n$ and $\dim(C'')=n-1$. 
  If $C^\#=C'\cap C''$ is a $(n-2)$-cube in $M$ then the sequence is correctly
  constructed. The rest of the case is like case 2.2.

  Case 2.4: $\dim(C')=\dim(C'')=n-1$. If $C^\#=C'\cap C''$ is a $(n-2)$-cube in $M$ 
  so the sequence is correctly constructed. The rest is analogue to the case 2.2.

  The algorithm terminates since in $\omega(s)$ contains only a finite number of 
  $n-1$- and $n$-cubes that have to be searched.

  It remains to show that the algorithm selects the cube $C''$ at some point of 
  time during its work.

\begin{Lemma}\label{kubus-existenz}
  Let $F=(C_0,\ldots,C_l)$ be the sequence we constructed in case 2 above and let $C_0$ the 
  cube $C'$ that contains the $n$-simplex $\sigma_1$.
  If none of the cases 2.1 to 2.4 is applicable then $C''$ is a element of $F$.
\end{Lemma}

\begin{Beweis}
  We assume for contradiction that $C''$ gets never in the sequence $F$.

  We define a set $c_F$ of $n$-cubes (of $\omega(s)$) to contain the cubes that 
  got marked or got in $F$ during the search process and that have the following property:
  Let $C_1,C_2$ be cubes in $c_F$ then is either $C_1\cap C_2$ a $(n-1)$-cube or there is 
  a sequence $C_1,\ldots,C_2$ such that the intersection of consecutive cubes are
  $(n-1)$-cubes.
  Let us denote the intersection of all those sets $c_F$ with $\c_F$.
  The set $\c_F$ is well-defined since it is constructed from $n$-cubes of set $\omega(s)$ 
  defined above. 
  Let $\overline\c_F$ be the set of $n$-cubes in $\omega(s)$ that are not in $F$ and let
  $\partial\c_F$ be the set of $(n-1)$-cubes that are subcubes of a cube in $\c_F$ and
  of a cube in $\overline\c_F$.

  By assumption $C''$ is not in $F$ therefore $C''$ cannot have been marked, since 
  all marked cubes were once in $F$.

  No $(n-1)$-cube $C$ with two $\beta$-components in $\COMP{M}$ can be in $\partial\c_F$ 
  since from the two $n$-cubes containing $C$ one had to be in $\c_F$, the other 
  in $\overline\c_F$. But the algorithm for case 2 would explore both $n$-cubes and so both 
  $n$-cubes end up in $F$ or get marked.

  Furthermore the points of $C_1\SETMINUS M$ and $C_2\SETMINUS M$ for two cubes 
  $C_1,C_2\in\partial\c_F$ are contained in the same $\beta$-component of $\omega(s)\SETMINUS M$
  if they exist. Let's try to see this:

  Case 1: A $(n-2)$-cube $C^*\SUBSET C_1\cap C_2$ exists. 
  
  Case 1.1:  $C^*\SUBSET M$. Then $C_1$ and $C_2$ are contained in $n$-cubes $C_1'$ 
  respectively $C_2'$ of $\overline\c_F$. Both $n$-cubes are contained in 
  $\bigcap_{p\in C^*}\omega(p)$ but not in different $\beta$-components of 
  $\omega(s)\SETMINUS M$, for otherwise one of the cubes or a common $(n-1)$-subcube
  would end up in $F$ and this contradicts their choice in $\overline\c_F$.

  Case 1.2: $C^*$ is not completely contained in $M$.
  The sets  $C_1\SETMINUS M$ and $C_2\SETMINUS M$ are $\beta$ connected since each of the
  cubes $C_1$ and $C_2$ can contain only one $\beta$-component in $\COMP{M}$ by the remark 
  above.

  Case 2: The intersection of $C_1$ and $C_2$ is no $(n-2)$-cube. By definition of
  $\c_F$ a sequence of $(n-1)$-cubes $G=(C_1,\ldots,C_2)$ exists such that consecutive
  elements share a common $(n-2)$-cube. This sequence can be obtained by restricting
  a sequence from $C_1'$ to $C_2'$ to $\partial\c_F$.
  We examine the case that in $G$ a $(n-1)$-cube $C_i\SUBSET M$ succeeds a $(n-1)$-cube
  $C_{i-1}$ that contains points of $\COMP{M}$. One of the $\beta$-components of
  $\bigcap_{p\in C_i}\omega(p)\SETMINUS M$ is $\beta$-connected to the $\beta$-component of
  $C_1$ in $\bigcap_{p\in C^*}\omega(p) \SETMINUS M$, if $C^*$ is the $(n-2)$-cube
  common to $C_i$ and $C_{i-1}$.
  If the two consecutive $(n-1)$-cubes $C_{i-1}$ and $C_i$ in $\partial\c_F$ are 
  contained in $M$, then we can also connect the $\beta$-components through 
  $\bigcap_{p\in C^*}\omega(p) \SETMINUS M$.
  If we iterate this argument, we can construct a $\beta$-path in $\COMP{M}$ from 
  $C_1\SETMINUS M$ to $C_2\SETMINUS M$ by using $G$. Refer to figure 
  \ref{Pic:drei_beta_komponenten}.

  One can see that $\c_F$ may contain $n$-cubes with two $\beta$-components in $\COMP{M}$.
  Some of them (maybe none) contain a subcube in $\partial\c_F$ that possesses points in 
  $\COMP{M}$ (see figure \ref{Pic:drei_beta_komponenten}). Therefore, we can define
  the $\beta$-component $K_1$ of $\bigcup\c_F\SETMINUS M$ as the one that has no points in 
  $\bigcup\partial\c_F$. The set $\bigcup\partial\c_F\SETMINUS M$ may be 
  empty, but if not, then $\bigcup\c_F\SETMINUS M$ contains two $\beta$-components.

  \begin{figure}[htb]
  \begin{center}
  \includegraphics{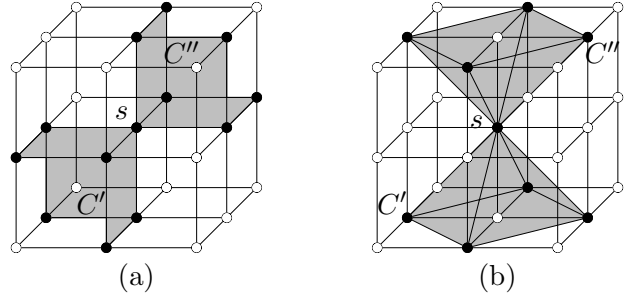}
  \end{center}
  \caption[Three $\beta$-components in $\omega(s)\SETMINUS
    M$]{\small{Two cases in 3D, such that the given algorithm would not find 
      the cube $C''$ by starting at $C'$. In both cases the set
      $\omega(s)\SETMINUS M$ has three $\beta$-components. In Figure (a) the set
      $\bigcup\partial\c_F\SETMINUS M$ is empty, in Figure (b) it is not.}} 
  \label{Pic:drei_beta_komponenten}
  \end{figure}

  Since $C''$ is no element of $F$, we can start the construction-algorithm for the 
  sequence $F$ with the cube $C''$ instead of $C'$. We get a sequence $F'$ that does not have
  $C'$ as an element, and we can define 
  $\c_{F'}$, $\partial\c_{F'}$ and $\overline\c_{F'}$ analog to the sets above.
  Thus, we get a $\beta$-component of $\bigcup\c_{F'}\SETMINUS M$ that has no points
  in $\bigcup\partial\c_{F'}\SETMINUS M$. Call this $\beta$-component $K_2$.
  The $\beta$-components $K_1$ and $K_2$ are necessarily distinct and not $\beta$-connected,
  since $\c_F\SUBSET\overline\c_{F'}$ and $\c_{F'}\SUBSET\overline\c_F$
  and every $\beta$-path between $K_1$ and $K_2$ would meet $\bigcup\partial\c_F$.
  The set $\bigcup\partial\c_F\SETMINUS M$ does not contain any points that are $\beta$-connected to 
  $K_1$. An analog argument holds for $K_2$.
  
  If the relation
  \begin{equation}
    \left(\bigcup\partial\c_F\right)\cap
    \left(\bigcup\partial\c_{F'}\right)\SETMINUS M\neq\emptyset
  \end{equation}
  holds, the set $\omega(s)\SETMINUS M$ has three $\beta$-components, since
  $\partial\c_F=\partial\c_{F'}$. This is not possible in a digital $(n-1)$-manifold
  because of property \ref{twocomponents}. Thus $C''$ would end up in $F$.

  If instead the relation  
  \begin{equation}
    \left(\bigcup\partial\c_F\right)\cap
    \left(\bigcup\partial\c_{F'}\right)\SETMINUS M=\emptyset
  \end{equation}
  holds, then $\partial\c_F=\partial\c_{F'}$ cannot be true, because any
  $(n-2)$-cube in $\partial\c_F$ is completely contained in $M$ and therefore it is also 
  contained in $F$ and in $F'$. This contradicts the assumption that $C''$ never gets 
  in $F$, because the algorithm would have discovered $C''$ eventually.

  Therefore, $\partial\c_F$ and $\partial\c_{F'}$ have to be distinct. 
  Since the sets $\c_F$ and $\c_{F'}$ are disjoint and $\bigcup\c_F\cap\bigcup\overline\c_F$ 
  is distinct from $\bigcup\c_{F'}\cap\bigcup\overline\c_{F'}$, it is possible to find 
  a $n$-cube $C$ in the set $\overline\c_F\cap\overline\c_{F'}$. This cube $C$ contains a point 
  $p$ of $\COMP{M}$ that is also in $\bigcup\c_F\cap\bigcup\c_{F'}$. The point $p$ cannot 
  be $\beta$-connected to $K_1$ or $K_2$. Again, three $\beta$-components do exist 
  in $\omega(s)\SETMINUS M$. 
  Again, the definition of a $(n-1)$-manifold is contradicted and so we have
  to conclude that $C''$ needs to end up in $F$.
\end{Beweis}

  The cube $C''$ is to be found by the algorithm and the sequence of simplices gets 
  constructed correctly between two cubes by Lemma \ref{hardercase}. This means in particular
  that the cube $D$ from case 2.1 will be discovered, if necessary.

  We now can use our knowledge how to construct sequences of $n$-simplices inside of cubes
  and between two cubes, to build a sequence of $n$-cubes from the arbitrary chosen
  points $p$ and $p'$ of $M$. Thereby the strong connectivity is proven.
\end{Beweis}

We see, that the simplicial complex $K'(M)$ is a $(n-1)$-pseudomanifold for
any given $(n-1)$-manifold, and we know that the geometric
realization $|K'(M)|$ of $K'(M)$ is a topological manifold in $\R^n$, i.e. a
polytope of dimension $n-1$ in $\R^n$. Thus, we can apply the original
Theorem of Jordan-Brouwer to show, that by Lemma \ref{glatt-diskret}
$M$ has the same topological properties as $|K'(M)|$. This means $M$
decomposes $\Z^n$ in exactly two connected $\beta$-components and is
itself the boundary of both. To see, that $M$ is the boundary of both
$\beta$-components, one has to show that no point can be removed from
$M$ without turning the set $\Z^n\SETMINUS M$ into one
$\beta$-component. This follows from the Property 2 of $M$.

\begin{Lemma}\label{simple-points}
  Let $M\SUBSET\Z^n$ be a $(n-1)$-manifold under the good pair
  $(\alpha,\beta)$.  Then $M$ contains no simple points. 
\end{Lemma}

\begin{Beweis}
  Any point $p\in M$ is $\beta$-adjacent to two $\beta$-components of
  $\omega(p)\SETMINUS M$. These $\beta$-components have a nonempty intersection
  with the two $\beta$-components of $\COMP{M}$ by Theorem \ref{satz-zwei-komp}.
  If we would remove $p$ from $M$ then those two $\beta$-components would become one.
  Thus $p$ cannot be simple.
\end{Beweis}

\section{Conclusions}

In this text we gave an overview on the study of the adjacency-relations
that admits the possibility of defining discrete analogs to
$(n-1)$-dimensional manifolds. To achieve this goal, we gave the
definition of a digital $(n-1)$-manifold. This definition is new in
the sense, that it generalizes to arbitrary dimensions and to
arbitrary good pairs of adjacency-relations. It relies heavily on the
concept of the separation property given in the definition \ref{trenndef}. 

During the research to this results, the author has shown that some known 
standard adjacency-relations give good pairs. The results will be published 
as soon as possible in another paper. Furtermore a set of axioms has been 
found that allows the formal definition of a digital geometry in arbitrary 
dimensions. This set of axioms is based on the work of Albrecht H\"ubler 
\cite{huebler}.

Further topics of research may include a definition of discrete
$k$-manifolds in $\Z^n$ for $0\le k\le n$. One also could try to
translate topological theorems from $\R^n$ to $\Z^n$.



\begin{thebibliography}{99}
\bibitem{alexandrov} P.S.~Alexandrov. \emph{Combinatorial
  Topology. Volumes 1,2 and 3.} Reprint, Dover Publications,
  Inc. Mineola, New York, 1998.
\bibitem{evako} A.V.Evako, R.Kopperman, Y.V.Mukhin. \emph{Dimensional
  Properties of Graphs and Digital Spaces.} Journal of Mathematical
  Imaging and Vision 6, 109-119 (1996).
\bibitem{herman} Gabor T. Herman, \emph{Geometry of Digital Spaces.}
  Birkhäuser Boston, 1998.
\bibitem{hermanzhao} Gabor T. Herman, Enping Zhao. \emph{Jordan
  Surfaces in Simply Connected Digital Spaces.} Journal of
  Mathematical Imaging and Vision 6, 121-138, 1996.
\bibitem{huebler} Albrecht H\"ubler. \emph{Diskrete Geometrie für                                                                                                                                            
  die digitale Bildverarbeitung.} Dissertationsschrift, FSU Jena,                                                                                                                                          
  1989.                                                                                                                                                                                                    
\bibitem{khalimsky} Efim Khalimsky. \emph{Topological structures in
  computer science.} J. Appl. Math. and Sim. Vol 1, 25-40, 1987.
\bibitem{khachan} Mohammed Khachan, Patrik Chenin, Hafsa
  Deddi. \emph{Digital pseudomanifolds, digital weakmanifolds and
    Jordan-Brouwer separation theorem.} Disc. Appl. Mathematics 125,
  45-57, 2003. 
\bibitem{kiselman} Christer O. Kiselman. \emph{Digital Jordan Curve
  Theorems.} Online Ausgabe: \verb|citeseer.ist.psu.edu/411748.html|, 2000.
\bibitem{klette} Reinhardt Klette, Azriel Rosenfeld. \emph{Digital
  Geometry.} Morgan Kaufmann Publishers, 2004.
\bibitem{kong} T. Y. Kong. \emph{Can 3-D Digital Topology be Based on
  Axio\-matic\-ally De\-fined Digi\-tal Spaces?} Online Version:\\
  \verb|http://citr.auckland.ac.nz/dgt/Problem_Files/GoodPairs.pdf|, 2001.  
\bibitem{kongrose} Azriel Rosenfeld, T.Y. Kong. \emph{Digital
  topology: introduction and survey.} Computer Vision Graphics and
  Image Processing 48 (3), 357-393, 1989.
\bibitem{rosenfeld} Azriel Rosenfeld. \emph{Arcs and Curves in digital
  pictures.} J. ACM 20, 81--87, 1973.
\end{thebibliography}
\end{document}